\documentclass{article}
\usepackage[T1]{fontenc}
\usepackage{iclr2027_conference,times}
\iclrfinalcopy
\usepackage{amsmath,amsfonts,bm}

\def\eqref#1{equation~\ref{#1}}
\def\1{\bm{1}}

\DeclareMathAlphabet{\mathsfit}{\encodingdefault}{\sfdefault}{m}{sl}
\SetMathAlphabet{\mathsfit}{bold}{\encodingdefault}{\sfdefault}{bx}{n}

\usepackage{booktabs,graphicx,xspace,longtable,pdflscape,float}
\usepackage[section]{placeins}
\usepackage{xcolor,colortbl}
\usepackage{hyperref,url}
\usepackage[most]{tcolorbox}
\usepackage{enumitem}
\hypersetup{pdflinkmargin=1.5pt}
\definecolor{TableLavender}{HTML}{F3F2FF}
\definecolor{TableApricot}{HTML}{FFF5E9}
\definecolor{TableMint}{HTML}{DDF8DF}
\definecolor{TableHeader}{HTML}{F3F3F3}

\newsavebox{\SummaryMeasureBox}
\newlength{\SummaryExtraPadding}
\NewDocumentEnvironment{summarytabular}{m m +b}{%
  \color{black}\bfseries
  \sbox{\SummaryMeasureBox}{\begin{tabular}{#2}#3\end{tabular}}%
  \setlength{\SummaryExtraPadding}{\dimexpr\linewidth-\wd\SummaryMeasureBox\relax}%
  \divide\SummaryExtraPadding by\numexpr2*#1\relax
  \addtolength{\tabcolsep}{\SummaryExtraPadding}%
  \begin{tabular}{#2}#3\end{tabular}%
}{}

\newcommand{\method}{AutoMIP\xspace}
\floatstyle{ruled}
\newfloat{algorithm}{tbp}{loa}
\floatname{algorithm}{Algorithm}
\newtcolorbox{promptbox}[1]{
  enhanced,
  breakable,
  colback=gray!4,
  colframe=gray!65,
  colbacktitle=black!75,
  coltitle=white,
  fonttitle=\bfseries,
  title={#1},
  attach boxed title to top left={
    xshift=0.35cm,
    yshift=-2mm
  },
  boxed title style={
    colback=black!75,
    colframe=black!75,
    boxrule=0pt,
    arc=1.5mm
  },
  arc=2mm,
  boxrule=0.8pt,
  left=3mm,
  right=3mm,
  top=3mm,
  bottom=2mm
}


\title{Autoresearch in Mixed-Integer Linear and Nonlinear Programming}
\author{
Yuwei Gu$^{1}$ \,
Yaoxin Wu$^{2*}$ \,
Tong Guo$^{3}$ \,
Wen Song$^{4}$ \,
Zhiguang Cao$^{5}$ \\
$^{1}$Chengdu University of Information Technology
\quad
$^{2}$Eindhoven University of Technology
\quad\\
$^{3}$Nanyang University of Technology
\,
$^{4}$Shandong University
\,
$^{5}$Singapore Management University
}
\begin{document}
\raggedbottom
\setlength{\textfloatsep}{10pt plus 2pt minus 2pt}
\setlength{\intextsep}{6pt plus 2pt minus 2pt}
\setlength{\floatsep}{8pt plus 2pt minus 2pt}
\maketitle

% \begin{abstract}
% Despite recent progress in autoresearch, applying it to practical operations research problems, which are typically formulated as NP-hard mixed-integer linear or nonlinear programs (MILPs or MINLPs), requires more systematic and efficient research capabilities to manage  competing ideas and experimental trajectories. We introduce \method, a reusable agent skill for organizing long-horizon autoresearch in MIPs through idea pooling and algorithm tree search.
% AutoMIP maintains a persistent pool of complementary candidate ideas, and organizes executable experiments in an algorithm tree, allowing the agent to preserve untried hypotheses, refine promising algorithms, and switch to new methodological directions from historical states.
% \method achieves the highest final success rate for MILP or MINLP benchmark cohorts. In MIPLib, AutoMIP reaches new best solutions for 31 out of 60 instances, more than current autoresearch frameworks; 
% In MINLPLib, AutoMIP is more advantageous with a success rate of 52 out of 60. 
% The complementary effects of idea pooling and algorithm tree search are evidenced by the ablation study.
% \end{abstract}

\begin{abstract}
Despite recent progress in autoresearch, applying it to practical operations research problems—typically formulated as NP-hard mixed-integer linear or nonlinear programs (MILPs or MINLPs)—remains challenging because effective research requires systematically managing competing ideas and long-horizon experimental trajectories. We introduce \method, a reusable agent skill for organizing long-horizon autoresearch in mixed-integer programming through idea pooling and algorithm tree search. \method maintains a persistent pool of complementary candidate ideas while organizing executable experiments into an algorithm tree, enabling the agent to preserve unexplored hypotheses, refine promising algorithms, and switch to alternative methodological directions based on historical states. On MILP and MINLP benchmark cohorts, \method achieves the highest final success rates among the evaluated autoresearch frameworks. On MIPLib, \method discovers new best solutions for 31 of 60 instances, surpassing existing autoresearch frameworks. On MINLPLib, it achieves new best solutions for 52 of 60 instances. Ablation studies further demonstrate the complementary contributions of idea pooling and algorithm tree search, highlighting the importance of jointly maintaining diverse research ideas and structured experimental trajectories for long-horizon autoresearch.
\end{abstract}

\section{Introduction}
Autoresearch agents improve algorithms by proposing changes, executing experiments, and using measured outcomes to decide what to try next~\citep{zheng2025automation}. Execution-guided discovery already demonstrates the value of feeding successful programs and heuristic ideas back into generation. However, existing autoresearch agents have primarily been developed for machine learning and scientific discovery, with applications spanning domains such as chemistry, biology, and materials science. NP-hard operations research (OR) problems modeled as mixed-integer linear or nonlinear programs (MILPs or MINLPs) have received comparatively less attention \citep{gridach2025agentic,ren2025towards}, which have broad applications across  domains such as manufacturing \citep{naderi2023mixed}, logistics \citep{zhou2023exact}, 
finance \citep{ararat2023computation}.

Given an OR problem formulated as a MILP or MINLP, a straightforward approach is to apply an exact solver, such as Gurobi or SCIP \citep{clautiaux2025last}. However, exact solvers may require substantial computational time on challenging instances. To obtain high-quality solutions more quickly, learning-guided approaches enhance different components of the optimization process, including primal heuristics \citep{han2023gnn,huang2024contrastive,liu2025apollo}, branching heuristics \citep{zhang2024towards,sun2024mgmatch,feng2025sorrel}, and cutting-plane selection \citep{tang2020reinforcement,wang2024learning,paulus2022learning}. Despite their promise, these approaches typically require substantial expertise in deep learning and considerable problem-specific model design and tuning to achieve effective performance on a given class of optimization problems.

With the advance of current large language models (LLMs), LLM-assisted algorithm design  has shown promise for automatically solving OR problems through heuristic design \citep{liu2024llm4ad,Funsearch}. By generating and iteratively evaluating heuristic designs, existing methods explore algorithmic spaces through population-based search \citep{huang2026automatic,eoh,reevo} or tree-based search \citep{mcts-ahd,wang2025planning,kiet2026motif}, motivating the automatic design of high-performing decomposition-enhanced HGS solvers for the LSCVRP \citep{guo2026automated}. Meanwhile, autoformulation work  applies LLMs to automate the problem modeling process
\cite{ramamonjison2023nl4opt,ahmaditeshnizi2024optimus,
xiao2024chain,huang2025orlm,jiang2025llmopt}. However, these approaches primarily focus on generating or refining individual algorithmic designs or formulations, without explicitly managing diverse research hypotheses together with their corresponding experimental states throughout an iterative research process. 

For NP-hard optimization, autoresearch is inherently stateful: a long research run accumulates reusable artifacts, including verified solutions, executable code, failure diagnoses, and untested hypotheses. These artifacts support different decisions: a strong incumbent can be retained while changing research direction; 
a failed experiment can motivate a repair; an untested hypothesis can become relevant as new evidence emerges. Consequently, each experiment involves two main coupled decisions: \emph{which research direction to pursue} and \emph{which historical state to use as its starting point}. Different directions may depend on different representations, algorithmic components, or previously developed code, making some states suitable starting points and others incompatible. Effective autoresearch therefore requires jointly managing research hypotheses and their compatible executable states, so that accumulated evidence can inform both \emph{what to explore} and \emph{where to start}.

In this paper, we propose a systematic framework, \method, for autoresearch on MILP or MINLP optimization. \method couples a persistent idea pool with an algorithm tree to jointly manage research hypotheses and executable experimental states. A review-and-ideation layer generates and maintains diverse research directions from accumulated evidence; an algorithm-search layer selects refinements or switches between directions while restoring compatible historical states and artifacts; and a verification layer executes candidates, validates their outcomes, and feeds the resulting evidence back into the exploration process. This design enables an LLM agent to iteratively deepen promising directions, revisit earlier states, and explore alternatives within an end-to-end research loop. We demonstrate \method  for solving challenging optimization problems in MILP or MINLP benchmark and show that it can discover effective improvements over the best known solutions under less computational budget than existing autoresearch frameworks. Our contributions are threefold: 
% \begin{itemize}
%     \item We propose \method, a systematic autoresearch framework for automated optimization algorithm development that couples a idea pool with an algorithm tree. The idea pool maintains diverse, executable research directions generated from the accumulated context and experimental evidence, while the algorithm tree records hypotheses, executable states, artifacts, and their dependencies, enabling the agent to either refine promising directions or switch to alternative ones throughout the research process.
%     \item We consider the next research direction and the historical experimental state from which to pursue it. When switching to a new direction, the agent identifies compatible previously explored states, restores the corresponding artifacts, and continues experimentation from the selected state, thereby reusing accumulated experimental knowledge rather than restarting from scratch
%     \item We integrate hypothesis generation, algorithm search, execution, and verification into an end-to-end autoresearch loop and evaluate \method on challenging MILP or MINLP problems under fixed computational budgets. The experiments demonstrate that \method aches new best
% solutions for 31 out of 60 instances, more than current autoresearch frameworks;
% In MINLPLib, AutoMIP is more advantageous with a success rate of 52 out of
% 60. 
% \end{itemize}
\begin{itemize}
\item We propose \method, a systematic autoresearch framework for  MILP or MINLP optimization that couples an idea pool with an algorithm tree. The idea pool maintains diverse research directions, while the algorithm tree records experimental states, artifacts, and their dependencies, throughout an iterative research
process.
\item During the autoresearch, we enable a joint selection of the next research direction and its associated starting state in algorithm tree. When switching research directions, \method identifies a compatible historical state, restores its artifacts, and continues experimentation from it, enabling effective reuse of accumulated knowledge.
\item We apply the end-to-end autoresearch loop and evaluate it \method on challenging MILP and MINLP benchmarks under fixed computational budgets. \method finds new best solutions for 31 of 60 instances in MIPLib, and 52 out of 60 on MINLPLib, outperforming existing autoresearch frameworks.
\end{itemize}

\section{Related Work}
We review important approaches in current autoresearch work, including experience reuse, alternative organization, experiment allocation, which are relevant to the main components in AutoMIP. 
\paragraph{Experience Reuse.}
 Recent approaches such as ExpeL \citep{zhao2024expel} and AgentKB \citep{tang2025agentkb} show that LLMs improve by reusing past trajectories, such as reflections or code edits from past attempts.
These methods have proven effective on reasoning tasks such as information retrieval and web browsing. Interleaving reasoning, action, and reflection enables agents to use execution feedback to adapt subsequent decisions \citep{react,reflexion}. Other approaches structure experience by decomposing tasks into subgoals and hierarchies, or by organizing and retrieving related memories through linked and hierarchical representations \citep{hiagent,tme,amem,raptor,memtree}. MemoryArena further evaluates whether agents can reuse experience across interdependent tasks \citep{memoryarena}.

\paragraph{Alternative Organization.}
% Explicit search complements memory by organizing which alternatives can be revisited, rather than only which observations can be retrieved. Tree of Thoughts evaluates intermediate reasoning paths and supports lookahead and backtracking \citep{tot}. LATS combines Monte Carlo tree search with language-model value estimates, reflection, and environmental feedback to guide reasoning and action \citep{lats}. Structured memory can also support this search: D-SMART performs tree-based reasoning over a dynamic dialogue knowledge graph \citep{dsmart}. These approaches make the choice of a continuation explicit. For long-running executable research, that choice also has an operational requirement: the agent must recover the code and artifacts on which the next experiment depends. A reusable research state must therefore preserve both decision evidence and the artifacts required for continuation.
Explicit search complements memory by structuring how alternative decisions are explored and revisited. Tree of Thoughts evaluates intermediate reasoning states to support lookahead and backtracking \citep{tot}, while LATS combines tree search, value estimation, reflection, and environmental feedback to guide subsequent actions \citep{lats}. D-SMART further organizes search over a dynamically evolving knowledge graph, enabling structured exploration of related reasoning paths \citep{dsmart}. These approaches make the selection and continuation of alternative paths explicit. For long-running research, however, continuation also requires recovering the code and artifacts associated with the selected path. Thus, an effective research state must preserve not only the evidence for choosing a continuation, but also the executable artifacts required to resume it.

\paragraph{Experiment Allocation.}
% Execution-guided evolution connects candidate generation to measured outcomes. FunSearch preserves diverse programs in an island-based population \citep{Funsearch}; EoH evolves natural-language heuristic descriptions together with executable code \citep{eoh}. In both cases, evaluated candidates provide material for subsequent generation, linking the search for new algorithms to the reuse of earlier solutions. This perspective motivates an important distinction in autoresearch: an evaluated program is an available implementation, whereas an untried hypothesis specifies an experiment that has yet to be constructed. Maintaining both kinds of information makes it possible to retain a promising alternative while continuing to refine an existing implementation. Experiment allocation then concerns which candidate to generate, when to leave the current direction, and which earlier work to reuse.
Execution-guided evolution connects candidate generation with measured experimental outcomes. FunSearch maintains diverse executable programs in an island-based population \citep{Funsearch}, while EoH \citep{eoh} jointly evolves natural-language heuristic ideas and their implementations, allowing successful candidates to seed subsequent generation. CORAL extends this paradigm to long-running multi-agent evolution through persistent shared memory and asynchronous experimentation \citep{qu2026coral}.  However, these methods primarily reuse evaluated implementations rather than explicitly preserving untested research hypotheses or their compatible experimental states. Consequently, experiment allocation remains centered on generating the next candidate, rather than jointly deciding what direction to explore and where to resume within a long-running research process.

% Long-horizon autoresearch brings memory, search, and experiment allocation together in a concrete control problem. A new hypothesis can require code or solver artifacts from an earlier experiment; applying it to the current branch can carry over assumptions that no longer fit. 
% \method focuses on the joint choice of an untried direction and a compatible starting state. Its Persistent Candidate Pool retains hypotheses independently of the active experiment, while its Branching Exploration Tree records refinement and switching decisions together with reusable parent states. The contribution is this coupling: selecting a direction prompts an explicit comparison of the states from which its experiment can be executed, rather than treating hypothesis selection and artifact reuse as separate decisions. 
In contrast, \method treats autoresearch as a joint search over research directions and executable states. Its persistent idea pool maintains diverse, untried hypotheses independently of the active experiment, while its algorithm Tree records refinement and switching trajectories together with reusable parent states and artifacts. Rather than selecting a hypothesis and reusing artifacts in separate stages, \method explicitly couples these decisions by identifying the compatible historical state from which each new direction can be executed.

\section{\method}
\label{sec:task}
% \method organizes autoresearch around two coupled pillars: the hypotheses (i.e., research direction) for the next experiment and the experimental state from which to execute it. We propose the idea pool which is persist to supply complementary research directions, and the algorithm tree preserves the experimental dependencies of research directions. Together, they let the autoresearch agent deepen a productive direction, pursue an alternative, and reuse earlier work throughout an end-to-end research process.

\method organizes autoresearch around two coupled pillars: the research direction for the next experiment and the experimental state from which to execute it. We propose a persistent idea pool that maintains complementary research directions, while the algorithm tree preserves the experimental dependencies among directions and their executable states. Together, they enable the autoresearch agent to deepen a productive direction, pursue alternatives, and reuse earlier work throughout an end-to-end research process.

\begin{figure}[t]
\centering
\includegraphics[width=\linewidth]{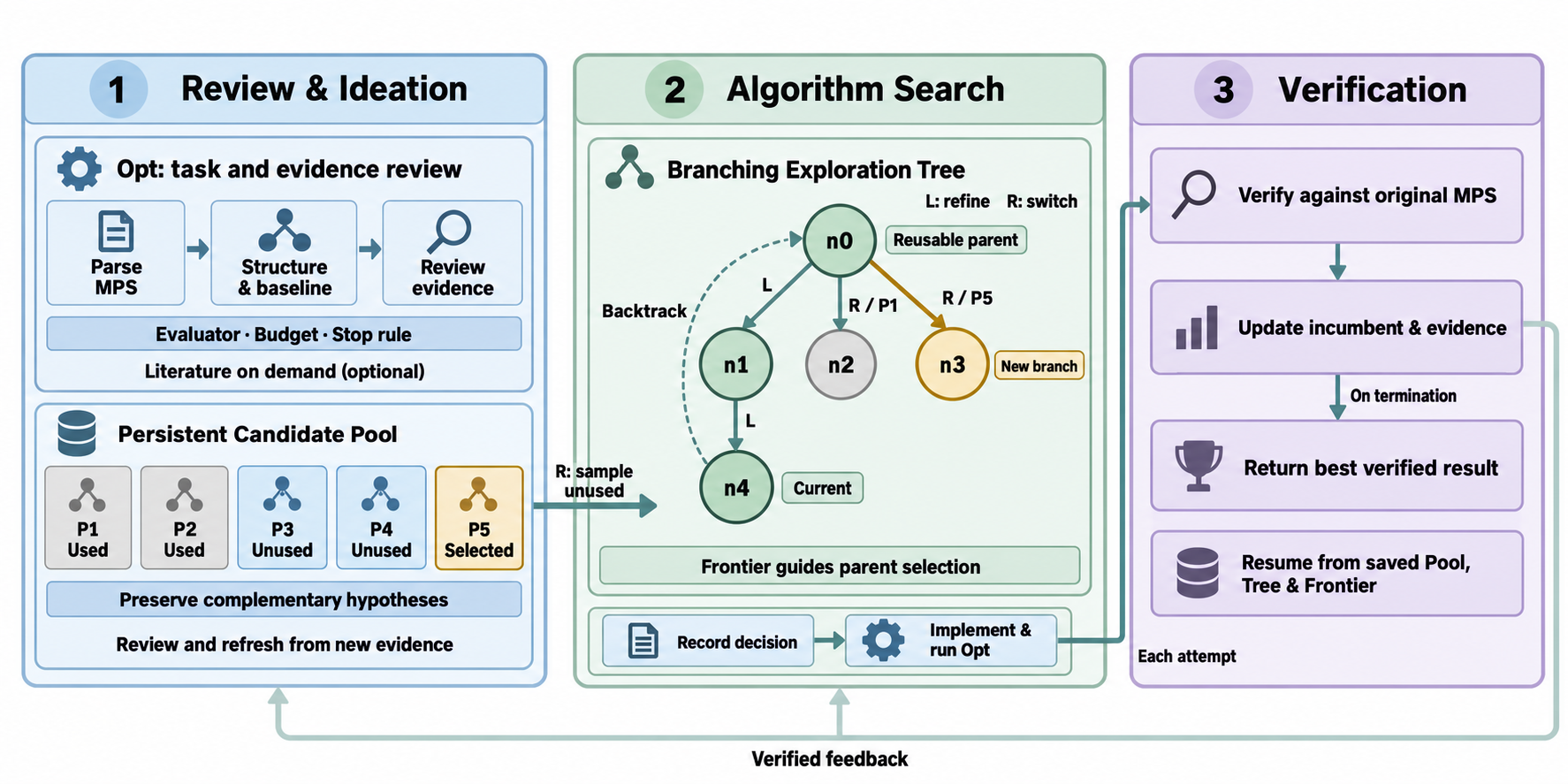}
\caption{\textbf{\method workflow.} 
% Opt denotes the primary task skill involving task analysis, candidate execution, and evaluation. 
\method preserves candidate hypotheses in idea pool, selects refinements or new branches (research directions) from reusable experimental states in algorithm tree, and updates idea pool and algorithm tree from experimental feedback. The workflow supports MILP and MINLP  optimization, with MPS taken as an example input format.}
\label{fig:loop}
\end{figure}

\subsection{Overview}
Within an LLM, the primary task skill defines the contract $\mathcal{C}=(\mathcal{S},E,B,\tau)$ representing a candidate snapshot interface
% (i.e., how executable candidate states are represented and restored) 
($\mathcal{S}$), evaluator ($E$), budget ($B$), and stopping predicate ($\tau$). The skill constructs, executes, and verifies candidates (see Appendix~\ref{app:opt-protocol} for skill details), while \method controls exploration with the skill. The exploration state $M_t=(T_t,F_t,P_t,c_t,A_t)$ contains the algorithm tree $T_t$, frontier nodes $F_t$, idea pool $P_t$, current node in algorithm tree $c_t$,
% \textcolor{red}{summaries}, 
and artifacts $A_t$ at step $t$, which is used throughout the reserch process. 

Figure~\ref{fig:loop} presents three layers of \method. \emph{Review-and-ideation} layer produce executable hypotheses according to the task and its evidence, resulting in a pool of ideas. \emph{Algorithm-search} layer selects a refinement or switch and an associated parent state, which is guided by algorithm tree. \emph{Verification} layer performs the experiment and evaluates the hypotheses and returns the result to the idea pool and algorithm tree. We elucidate the three layers in \method in next sections.

% The following layers specify the information maintained at each stage and how it passes to the next.

\subsection{Review-and-Ideation Layer}
\label{sec:grounding}
% This layer turns task evidence into a structured agenda for exploration. It comprises evidence review, candidate generation and screening, pool registration, and lifecycle maintenance. The review reads the task contract and goal, baseline and best verified result, structural constraints, execution logs, recurring failures, and saved artifacts. PoolThink then consolidates this evidence into candidate hypotheses and records why each direction is distinct and executable. The layer outputs a review record and a Persistent Candidate Pool $P_t$ for the search layer; it does not select a parent state or execute an experiment. Separating this agenda from the active branch keeps alternatives available while the current direction is refined.
The review-and-ideation layer converts the task and evidence into a structured set of research directions for subsequent exploration. 
This layer comprises three steps: evidence review, candidate generation and screening, pool collection. The review considers the following evidence: task contract and objective, the baseline and best verified results, structural constraints, execution logs, recurring failure modes, and previously generated artifacts. The agent is allowed to review literature in the very beginning without generated results, logs and artifacts. \emph{PoolThink} synthesizes this evidence into executable research directions, summarizes the rationale and requirements of each direction, and screens redundant or infeasible proposals. The layer outputs a pool of records (i.e., the idea pool)
% review records 
and maintains the pool $P_t$ for the algorithm-search layer. It does not determine a parent state in the algorithm tree or execute experiments. As such, separating candidate generation from active search preserves alternative directions while the current direction is being refined.

\textbf{Idea Pooling.} At the beginning of a problem-solving run, \emph{PoolThink} synthesizes evidence from task structure, current code, execution logs, and previous attempts to generate and screen five to ten complementary research directions.
% , screens proposed directions, and registers five to ten complementary candidates.
% Candidates differ in core hypotheses or strategy families, such as neighborhood search, structural decomposition, and model strengthening; parameter changes within one strategy remain local refinements.
Candidates are encouraged to differ in their core hypotheses or strategy families, such as neighborhood search, structural decomposition, or model strengthening, whereas parameter changes within the same hypotheses or strategy are treated as local refinements rather than separate directions. The pool is a generation-indexed collection of candidate records. Each record contains an identity, hypothesis (i.e., research direction), supporting evidence, difference from previously attempted directions, starting requirements, potential failure risks, and status (e.g., unused or selected). 
% These fields define the experiment, its artifact dependencies, and the outcome to inspect during verification. 
These fields characterize both the intended experiment and the artifacts it may require, i.e., the reusable executable outputs (e.g., source code, solver configurations, generated models, and intermediate files) associated with each experimental state.
At step $t$, let $U_t$ be the unused subset in the pool $P_t$:
\begin{equation}
q_t\sim\operatorname{Uniform}(U_t),\qquad
U_t=\{q\in P_t:\operatorname{status}(q)=\texttt{unused}\}.
\label{eq:pool-selection}
\end{equation}
% For a justified switch, the search layer samples $q_t$ uniformly from $U_t$. Registering its branch consumes the identity within that generation, while later $L$ attempts may refine the selected direction. Verification updates status and evidence, and the persistent pool is read again at subsequent solving steps and after interruptions. After at least five candidates have been used or completed, PoolThink may refresh the pool from accumulated evidence while preserving the recorded history. This reuse occurs within one instance-solving run; a new problem instance starts with a new task-specific review rather than carrying candidate assumptions over unchanged. Appendix~\ref{app:pool} gives the complete lifecycle rules.
When the algorithm-search layer initiates a direction switch, it samples $q_t$ uniformly from $U_t$. Registering the corresponding branch in the algorithm tree consumes the record. In other words, once a candidate direction is selected to create a new branch in algorithm tree, it is marked as \texttt{used} and cannot be selected again for another direction switch, while subsequent refinement steps may continue developing this selected direction into other candidates.
% , while subsequent refinement steps may continue developing the selected direction. 
Verification then updates its status and accumulated evidence, which remain available to later research decisions. 
% and after interruptions. 
Once at least five candidates have been used or completed, \emph{PoolThink} may refresh the pool using the accumulated evidence  while preserving the existing history. This persistence is maintained within a problem-solving run; a new problem instance begins with a new task-specific evidence review rather than inheriting candidate assumptions unchanged. Appendix~\ref{app:pool} provides the complete pool lifecycle.

\begin{figure}[t]
\centering
\includegraphics[width=0.68\linewidth]{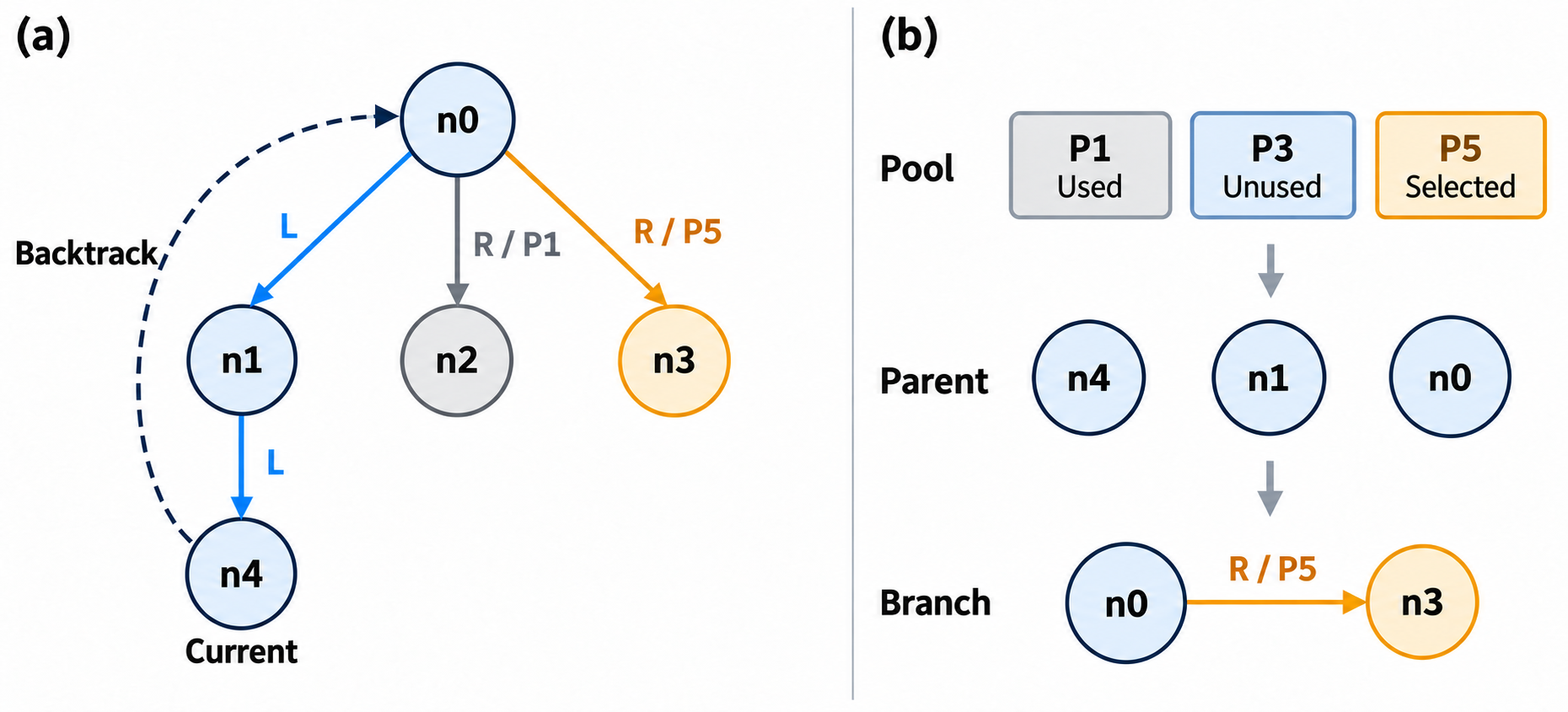}
\caption{Coupling a direction to its starting state. After n0$\rightarrow$n1$\rightarrow$n4, selecting P5 and reusing n0 creates the $R$/P5 branch n3. Solid edges record ancestry; dashed arc restores historical artifacts.}
\label{fig:tree-search}
\end{figure}

\subsection{Algorithm Search Layer}
\label{sec:exploration-loop}

The algorithm-search layer converts a research direction into an executable experiment. Its inputs are the persistent idea pool $P_t$, algorithm tree $T_t$, frontier nodes $F_t$, current node $c_t$ in the tree, and artifacts $A_t$. The layer determines whether to \emph{refine} the current research direction or \emph{switch} to an alternative one, identifies the appropriate historical starting state, restores its artifacts, and registers the experiment before execution. The measured outcome is produced later by the verification layer.

\textbf{Tree-Guided Algorithm Search.}
Algorithm Tree is the search structure for executable experimental states. Let $T_t=(V_t,D_t)$, where each node stores a research direction, observations, and reusable artifacts, and each directed edge records how the child experiment is derived from its parent. The tree is multiway, with two edge types: $L$ for local refinement and $R$ for switching to a different research direction. Algorithm~\ref{alg:pooltree} summarizes the search procedure, and Appendix~\ref{app:tree-protocol} provides the complete tree-search and branching-state protocol.

\begin{algorithm}[H]
% \caption{AutoMIP exploration with a primary task skill}
\caption{Algorithm Tree Search}
\label{alg:pooltree}
\small
\begin{tabbing}
\quad\=\quad\=\quad\=\kill
\textbf{Input:} Exploration state $M_t=(T_t,F_t,P_t,c_t,A_t)$.\\
\textbf{Output:} Registered experiment and restored experimental state.\\
Read the current node $c_t$, frontier nodes $F_t$, candidate pool $P_t$, and artifacts $A_t$.\\
\textbf{if} An admissible refinement exists:\\
\>Set $d_t=L$ and $p_t=c_t$.\\
\textbf{else if} an unused candidate is available:\\
\>Sample $q_t\sim\mathrm{Uniform}(U_t)$.\\
\>Construct compatible parent set $C_t(q_t)$.\\
\>Select $p_t\in C_t(q_t)$ using selection policy (Appendix~\ref{app:policy}).\\
\>Set $d_t=R$ and associate the experiment with $q_t$.\\
\textbf{else}: Terminate the search step.\\
Register $(p_t,v_{t+1},d_t)$ and restore artifacts of $p_t$.\\
\textbf{return} Experiment $v_{t+1}$ and restored  state.
\end{tabbing}
\end{algorithm}

The search decision is represented by $d_t\in\{L,R\}$, where an $L$ edge continues improving the current research direction, whereas an $R$ edge activates an unused research direction sampled from the idea pool $P_t$. Switching is triggered when the current research direction stagnates (without performance gain), its underlying assumption is refuted, or no further admissible refinement remains.

The layer determines not only \emph{what} to explore, but also \emph{where} to resume. 
% The frontier $F_t$ maintains executable historical states that remain available for continuation. 
% The frontier $F_t$ is the subset of nodes in the tree that can be expanded. It contains the current node and reusable historical nodes whose artifacts are available for continuation.
The frontier $F_t$ is the subset of reusable historical nodes whose artifacts are available for research continuation.
% Let $H_t$ denote the candidate historical states, including the current node, its ancestors, and reusable frontier nodes. 
Let $H_t$ denote the candidate parent nodes, including the current node, its ancestry, and reusable nodes in $F_t$.
A node $v$ is compatible with research direction $q$ if its saved artifacts satisfy the starting requirements of $q$, denoted by $K_t(v,q)=1$. The compatible parent set is
\begin{equation}
\begin{aligned}
C_t(q_t)&=\{v\in H_t:K_t(v,q_t)=1\},\\
p_t&=
\begin{cases}
c_t, & d_t=L,\\
% \arg\max_{v\in C_t(q_t)}\text{Policy}(v,q_t), & d_t=R.
p_t\in C_t(q_t), & d_t=R.
\end{cases}
\end{aligned}
\label{eq:tree-parent}
\end{equation}

The selected parent determines the executable state of the new experiment. For an example in Figure~\ref{fig:tree-search}, a new research direction may branch from an earlier node, allowing previously developed code and artifacts to be reused. 
% without inheriting incompatible intermediate modifications.
To prevent repeated switching without meaningful progress, we also track consecutive $R$ transitions. Let $r(v)$ denote the number of consecutive switching edges ending at node $v$, $\pi(v)$ denote its parent, and $\ell(v)$ its incoming edge, we define
\begin{equation}
r(v)=
\begin{cases}
0, & v\text{ is the root or }\ell(v)=L,\\
r(\pi(v))+1, & \ell(v)=R.
\end{cases}
\label{eq:right-chain}
\end{equation}

When repeated switches occur, the policy prioritizes compatible historical states over extending the latest branch. The complete selection policy is provided in Appendix~\ref{app:policy}. Finally, the layer registers the experiment in the algorithm tree by $V_{t+1}=V_t\cup\{v_{t+1}\},
D_{t+1}=D_t\cup\{(p_t,v_{t+1},d_t)\},$
% \begin{equation}
% V_{t+1}=V_t\cup\{v\},\qquad
% D_{t+1}=D_t\cup\{(p_t,v,d_t)\},
% \label{eq:tree-update}
% \end{equation}
where the new node $v$ stores its research direction $q$, parent state $\pi(v)$, branch rationale, and restored artifacts (i.e., the experimental state). 
% Recording the experiment before execution ensures that its lineage and reusable state are preserved even if the execution is interrupted.

% \subsection{Verification Layer}
% \label{sec:feedback}
% This layer turns an attempted algorithm into verified evidence. From the restored parent artifacts, the primary task skill constructs and executes the candidate, while evaluator $E$ checks feasibility, objective value, and task-specific acceptance conditions. It returns a validation result, updated exploration state, and the best verified result.

% The layer saves observations, validation evidence, and artifacts in the registered node, then updates the idea pool lifecycle, frontier, current pointer, and summaries. Success can support another $L$ refinement; diagnostic failure can identify a repair; repeated stagnation or refutation can justify an $R$ decision. The best verified result remains independent of the active branch, and persistent records support recovery after interruption. Execution stops when budget $B$ is exhausted or predicate $\tau$ is satisfied. Appendix~\ref{app:implementation} specifies the full update and recovery protocol.

\subsection{Verification Layer}
\label{sec:feedback}

The verification layer executes the registered experiment and converts its outcome into reusable research evidence. Starting from the restored experimental state, the primary task skill constructs and executes the candidate, while the evaluator  assesses feasibility, objective value, runtime, and other task-specific acceptance criteria. The layer outputs the evaluation result, updated exploration state, and the current best verified solution.

The verification result is then written back to the exploration state. The registered node stores its observations, evaluation evidence, and newly generated artifacts; the idea pool updates its lifecycle of the corresponding research direction; and the frontier and current node are refreshed for the next search step. A successful experiment may trigger further $L$ refinements, whereas repeated stagnation or refuted assumptions can justify a $R$ direction switch in algorithm tree. The research terminates when the computational budget $B$ is exhausted or the stopping predicate $\tau$ is satisfied. Appendix~\ref{app:verification-protocol} describes the complete verification, state-update, and recovery protocol.

\section{Experiments}
\label{sec:eval}
% Our experiments examine whether \method completes tasks earlier, succeeds on more instances, achieves the lowest final objective on more instances, and benefits from both exploration components. We evaluate three optimization cohorts, followed by component and agent-interface comparisons.

% \subsection{Experimental Setup}
% \label{sec:benchmarks-baselines-metrics}
% \paragraph{Tasks.}
We first assess AutoMIP in MIPLib. Open and Hard set each comprises 60 instances \citep{miplib}. For Open set, we seek a strict improvement over the best known solution,
% the starting incumbent, 
or first feasibility when no feasible solution is found yet, within 43,200 s. For Hard set, we seek a specified historical objective target within 3,600 s. We also use a 60-instance MINLPLib cohort with a 43,200 s horizon. Given these different tasks, we assess AutoMIP in terms of incumbent improvement, fixed-target attainment, and nonlinear optimization, respectively. 
% Component and interface comparisons use 20 Open instance identities. 
% The full-method ablation row reuses the corresponding Open observations; the interface comparison has its own runs. 
All configurations run on an Apple M4 Pro machine with 12 logical CPUs, 12 physical CPU cores, and 24~GB memory. Codex-based configurations use GPT-5.5, whereas the Claude Code interface uses Claude Sonnet 5. 
% The implementation of \method is available in the anonymous repository at \url{https://anonymous.4open.science/r/AutoMIP-1E22/}.

% \paragraph{Baselines.} \textbf{Codex} is the coding agent used directly for task analysis, code editing, execution, and verification \citep{codexcli}. \textbf{Loop} adapts the autoresearch cycle to optimization: modify a candidate, evaluate it, retain improvements or restore the retained state, and repeat \citep{autoresearch}. \textbf{AutoEoH} uses a coding agent to construct the task and evaluator for Evolution of Heuristics, which evolves heuristic descriptions and executable code from evaluation feedback \citep{eoh}. \textbf{EvoX} uses the evolutionary-computation framework to search candidate decisions or strategy parameters, with task solvers providing evaluation, repair, and verification \citep{evox}. Together these comparators cover direct agent use, sequential improvement, heuristic-code evolution, and population-based search. \textbf{Codex+\method (ours)} instantiates \method with the Codex coding agent, adding the proposed persistent candidate pool and branching exploration tree to the coding-agent workflow.

\paragraph{Baselines.}
\textbf{Codex} is the coding agent used directly for task analysis, code modification, execution, and verification \citep{codexcli}. \textbf{Loop} adapts the autoresearch cycle to optimization by iteratively modifying a candidate, evaluating its performance, retaining improvements or reverting to the best retained state, and continuing the search \citep{autoresearch}. \textbf{AutoEoH} employs a coding agent to construct the task and evaluation interface for EoH, which evolves heuristic descriptions and executable implementations based on evaluation feedback \citep{eoh}. \textbf{EvoX} uses an evolutionary-computation framework to search over candidate decisions or strategy parameters, with task-specific solvers providing evaluation, repair, and verification \citep{evox}. Together, these baselines represent four complementary paradigms: direct coding-agent use, sequential iterative improvement, heuristic-code evolution, and population-based search. We instantiates \textbf{\method} with Codex agent, augmenting the standard coding-agent workflow with our framework for  managing alternative research directions and reusable experimental states.

\paragraph{Metrics.}
We report Success to represent if the task objective is attained within the time horizon. We also report a win/solved count pair, runtime, and token consumption for each method in disjoint time intervals, followed by their average values. For attaining the historical objective target in Hard set, we only report the solved counts due to the fixed-target attainment target; all methods attain the same objective value whenever they solve an instance. Specifically, win records an instance on which a method achieves the lowest final objective value among all methods. Thus, win measures final solution quality, solved measures task attainment, and runtime measures completion speed. 
% Win comparisons involve the five methods on Open and MINLPLib, the three ablation variants, or the two agent-interface configurations, respectively. 
% The terminal 6--12 h interval includes runs still unfinished at the 12-hour horizon, so it can report time and token values even when its solved count is zero. 
Token values represent average token consumption over all instances (in each time interval), including unsuccessful runs. 
% token reductions compare the supplied Total-column means. 
The cutoffs are 3,600, 10,800, 21,600, and 43,200 s for the time intervals, except for Hard set, which uses 600, 1,200, 1,800, and 3,600 s. We set only one-hour time limit for Hard set because of its easier fixed-target attainment task and shorter completion times across methods. For method $m$, let $s_{im}$ indicate success and $T_{im}$ its time on instance $i$. 
% The cumulative coverage is defined by 
% \begin{equation}
% C_m(t)=\frac{1}{N}\sum_{i=1}^{N}\mathbf{1}\{s_{im}=1,\;0<T_{im}\leq t\},\qquad 0\leq t\leq H,
% \label{eq:coverage}
% \end{equation}
% where $N$ includes all cohort instances and $H$ is the horizon. Paired timing compares instances completed by both methods, using the median ratio $T_{i,b}/T_{i,\mathrm{PoolTree}}$ for comparator $b$; values above one favor \method. Average coverage is the time average of cumulative coverage over the full cohort horizon.

\subsection{Results on MIPLib}
\label{sec:miplib-results}
\paragraph{Open set.}
As presented in Table~\ref{tab:open}, \method completes 24 tasks within 3,600 s, compared with 13 for Codex, 15 for Loop, 12 for AutoEoH, and 6 for EvoX. 
% The gap over Codex is 18.3 percentage points. 
By three hours, \method has completed 29 tasks, matching Codex's final twelve-hour count. In total, AutoMIP finally achieves the goal (achieving the new best-known solutions) for 31 instances, more than the other methods. Moreover, most of its successful instances are achieved early within six hours, thus using shorter average time over all instances.
% to be useful under substantially shorter research windows.

\begin{table}[!htbp]
\caption{Open cohort, 60 instances.}
\label{tab:open}
\centering
\begingroup\fontsize{7}{8}\selectfont\setlength{\tabcolsep}{0.25pt}\renewcommand{\arraystretch}{2}
\begin{summarytabular}{16}{c*{3}{>{\columncolor{TableLavender}}c}*{3}{>{\columncolor{TableApricot}}c}*{3}{>{\columncolor{TableLavender}}c}*{3}{>{\columncolor{TableApricot}}c}*{3}{>{\columncolor{TableMint}}c}}\toprule
\multicolumn{1}{c}{\cellcolor{TableHeader}Method} & \multicolumn{3}{c}{\cellcolor{TableLavender}0--1 h} & \multicolumn{3}{c}{\cellcolor{TableApricot}1--3 h} & \multicolumn{3}{c}{\cellcolor{TableLavender}3--6 h} & \multicolumn{3}{c}{\cellcolor{TableApricot}6--12 h} & \multicolumn{3}{c}{\cellcolor{TableMint}Total} \\
\cmidrule(lr){2-4}\cmidrule(lr){5-7}\cmidrule(lr){8-10}\cmidrule(lr){11-13}\cmidrule(lr){14-16}
\multicolumn{1}{c}{\cellcolor{TableHeader}} & \multicolumn{1}{c}{\cellcolor{TableHeader}w/s} & \multicolumn{1}{c}{\cellcolor{TableHeader}time (s)} & \multicolumn{1}{c}{\cellcolor{TableHeader}token} & \multicolumn{1}{c}{\cellcolor{TableHeader}w/s} & \multicolumn{1}{c}{\cellcolor{TableHeader}time (s)} & \multicolumn{1}{c}{\cellcolor{TableHeader}token} & \multicolumn{1}{c}{\cellcolor{TableHeader}w/s} & \multicolumn{1}{c}{\cellcolor{TableHeader}time (s)} & \multicolumn{1}{c}{\cellcolor{TableHeader}token} & \multicolumn{1}{c}{\cellcolor{TableHeader}w/s} & \multicolumn{1}{c}{\cellcolor{TableHeader}time (s)} & \multicolumn{1}{c}{\cellcolor{TableHeader}token} & \multicolumn{1}{c}{\cellcolor{TableHeader}w/s} & \multicolumn{1}{c}{\cellcolor{TableHeader}time (s)} & \multicolumn{1}{c}{\cellcolor{TableHeader}token} \\ \midrule
\shortstack{AutoMIP} & 14/24 & 1,211 & 320,244 & 4/5 & 5,545 & 912,789 & 0/2 & 11,256 & 1,960,932 & 0/0 & 43,200 & 961,095 & 18/31 & 22,202 & 734,057 \\
Codex & 1/13 & 1,822 & 393,785 & 3/11 & 6,426 & 1,070,770 & 0/4 & 13,980 & 2,857,174 & 1/1 & 42,049 & 1,137,639 & 5/29 & 24,931 & 1,078,847 \\
Loop & 3/15 & 1,888 & 323,983 & 1/6 & 7,123 & 648,990 & 0/3 & 11,665 & 1,685,786 & 0/1 & 42,679 & 1,700,813 & 4/25 & 27,375 & 1,310,672 \\
AutoEoH & 0/12 & 2,157 & 606,353 & 3/9 & 7,142 & 1,035,190 & 0/4 & 14,175 & 2,556,814 & 0/1 & 43,162 & 2,660,954 & 3/26 & 27,626 & 1,999,227 \\
EvoX & 0/6 & 1,302 & 306,763 & 0/5 & 7,032 & 720,454 & 0/3 & 13,390 & 1,586,613 & 1/1 & 42,975 & 1,244,128 & 1/15 & 34,333 & 2,297,150 \\
\bottomrule\end{summarytabular}
\endgroup

\end{table}

For 28 Open instances completed within budget by both \method and Codex, \method finishes earlier in 24 instances (85.7\%), with an average speedup $\times$2.18.
It is also faster on 19 of the 23 instances jointly solved by \method and Loop, and on 22 of the 24 instances jointly solved by \method and AutoEoH, with an average speedup $\times$1.74 and $\times$3.09, respectively. In terms of final solution quality, \method achieves the best objective on 18 instances, compared with 5 for Codex, 4 for Loop, 3 for AutoEoH, and 1 for EvoX. Thus, on the Open instances, \method shows gains in both the speed of reaching its final solution and the quality of the final objective.

% It also finishes earlier on 19 of 23 common Loop successes and 22 of 24 common AutoEoH successes, with median ratios of 1.74 and 3.09. Final solution quality also favors \method: it records 18 objective wins, compared with 5 for Codex, 4 for Loop, 3 for AutoEoH, and 1 for EvoX. The Open gains therefore extend beyond reaching an improvement sooner to obtaining the best final objective on more instances.

\paragraph{Hard set.}
Comparison in terms of fixed-target attainment shows a similar pattern under a one-hour budget. In Table~\ref{tab:hard}, \method reaches 14 targets within 600 s versus 6 for Codex, and 27 within 1,800 s versus 17. Halfway through the budget, AutoMIP has already obtained 90.0\% of its eventual 30 successes. Codex can only reaches 29 till the end. 
% For a thirty-minute deadline, the earlier completions provide ten additional instances with a verified target already attained.

\begin{table}[!htbp]
\caption{Hard cohort, 60 instances and a one-hour horizon.}
\label{tab:hard}
\centering
\begingroup\fontsize{7}{8}\selectfont\setlength{\tabcolsep}{0.25pt}\renewcommand{\arraystretch}{2}
\begin{summarytabular}{16}{c*{3}{>{\columncolor{TableLavender}}c}*{3}{>{\columncolor{TableApricot}}c}*{3}{>{\columncolor{TableLavender}}c}*{3}{>{\columncolor{TableApricot}}c}*{3}{>{\columncolor{TableMint}}c}}\toprule
\multicolumn{1}{c}{\cellcolor{TableHeader}Method} & \multicolumn{3}{c}{\cellcolor{TableLavender}0--10 min} & \multicolumn{3}{c}{\cellcolor{TableApricot}10--20 min} & \multicolumn{3}{c}{\cellcolor{TableLavender}20--30 min} & \multicolumn{3}{c}{\cellcolor{TableApricot}30--60 min} & \multicolumn{3}{c}{\cellcolor{TableMint}Total} \\
\cmidrule(lr){2-4}\cmidrule(lr){5-7}\cmidrule(lr){8-10}\cmidrule(lr){11-13}\cmidrule(lr){14-16}
\multicolumn{1}{c}{\cellcolor{TableHeader}} & \multicolumn{1}{c}{\cellcolor{TableHeader}solved} & \multicolumn{1}{c}{\cellcolor{TableHeader}time (s)} & \multicolumn{1}{c}{\cellcolor{TableHeader}token} & \multicolumn{1}{c}{\cellcolor{TableHeader}solved} & \multicolumn{1}{c}{\cellcolor{TableHeader}time (s)} & \multicolumn{1}{c}{\cellcolor{TableHeader}token} & \multicolumn{1}{c}{\cellcolor{TableHeader}solved} & \multicolumn{1}{c}{\cellcolor{TableHeader}time (s)} & \multicolumn{1}{c}{\cellcolor{TableHeader}token} & \multicolumn{1}{c}{\cellcolor{TableHeader}solved} & \multicolumn{1}{c}{\cellcolor{TableHeader}time (s)} & \multicolumn{1}{c}{\cellcolor{TableHeader}token} & \multicolumn{1}{c}{\cellcolor{TableHeader}solved} & \multicolumn{1}{c}{\cellcolor{TableHeader}time (s)} & \multicolumn{1}{c}{\cellcolor{TableHeader}token} \\ \midrule
\shortstack{AutoMIP} & 14 & 399 & 96,926 & 8 & 869 & 120,307 & 5 & 1,511 & 154,571 & 3 & 3,454 & 171,106 & 30 & 2,035 & 145,646 \\
Codex & 6 & 322 & 196,416 & 6 & 909 & 116,999 & 5 & 1,471 & 153,526 & 12 & 2,912 & 206,999 & 29 & 2,333 & 192,485 \\
Loop & 4 & 518 & 80,309 & 8 & 873 & 102,209 & 4 & 1,364 & 135,665 & 6 & 3,218 & 228,227 & 22 & 2,602 & 195,393 \\
AutoEoH & 2 & 478 & 97,654 & 6 & 923 & 110,134 & 4 & 1,471 & 142,942 & 7 & 2,943 & 182,870 & 19 & 2,561 & 170,094 \\
EvoX & 3 & 574 & 117,775 & 6 & 961 & 101,781 & 4 & 1,546 & 203,375 & 2 & 2,894 & 254,562 & 15 & 2,495 & 229,032 \\
\bottomrule\end{summarytabular}
\endgroup

\end{table}

\begin{figure}[!htb]
\centering\includegraphics[width=\linewidth]{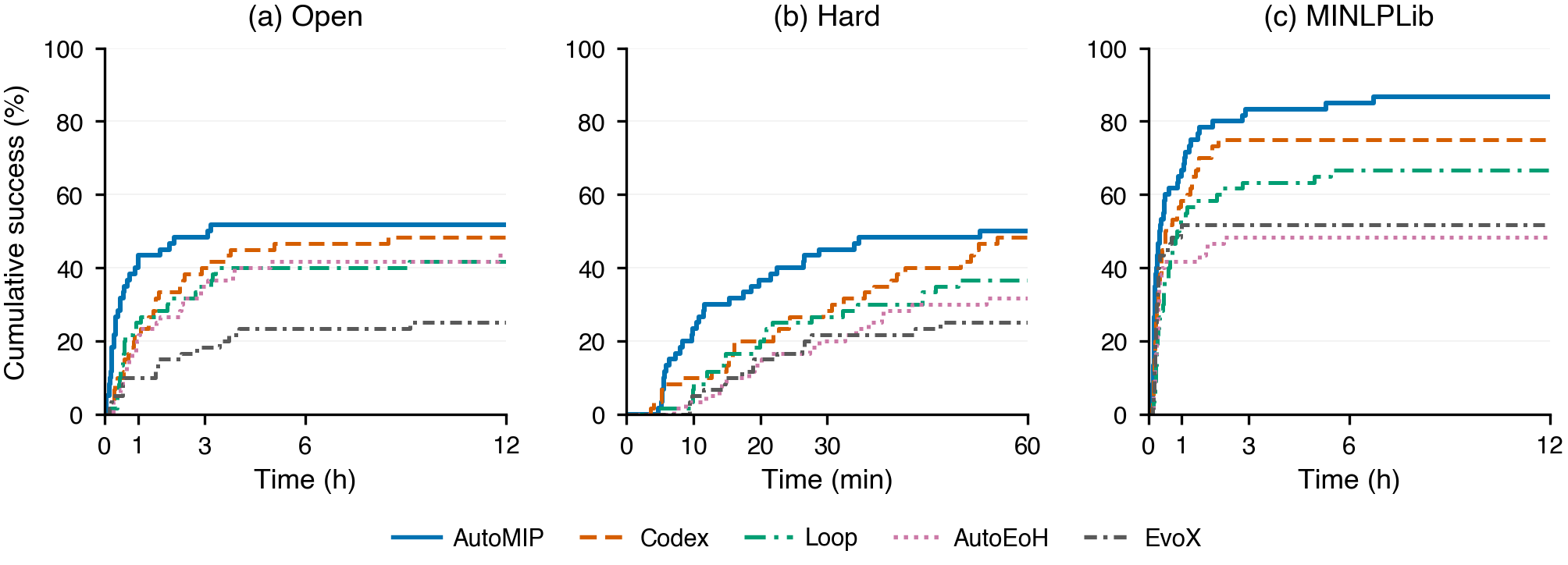}
\caption{Cumulative task success from per-instance timestamps. Each step adds a successful instance; the denominator is 60 in every panel. Curves extend to 43,200 s for Open and MINLPLib and 3,600 s for Hard. The cutoffs agree with Tables~\ref{tab:open}, \ref{tab:hard}, and \ref{tab:minlplib}, respectively.}
\label{fig:curve}
\end{figure}

Among 27 Hard instances completed within budget by both \method and Codex, \method is faster in 20 cases, with an average speedup $\times$2.00. 
% with a median time ratio of 2.00. 
The speedup against Loop and AutoEoH is $\times$2.61 and $\times$2.02. Figure~\ref{fig:curve}(a,b) shows the resulting shift toward earlier completion under both the improvement and fixed-target objectives. 
% Average coverage over the full horizon is 48.6\% versus 41.4\% for Codex on Open and 37.3\% versus 27.1\% on Hard.

Notably, earlier completion of AutoMIP is accompanied by lower token usage. In Tables~\ref{tab:open} and~\ref{tab:hard}, \method uses total 734,057 versus 1,078,847 tokens for Codex for Open set (32.0\% lower), and 145,646 versus 192,485 on Hard set (24.3\% lower).

% \FloatBarrier
\subsection{Ablation Study}
\label{sec:ablation-study}
% NoPool retains the algorithm tree that generates a research direction when a switch is needed. NoTree retains the persistent idea pool and random selection of unused candidates, while recording attempts in a linear sequence. Appendix~\ref{app:ablation-definitions} specifies the retained and removed operations.
NoPool removes the idea pool while retaining algorithm tree, generating a new research direction only when a direction switch is triggered. NoTree retains the idea pool and its random selection of unused research directions, but replaces the algorithm tree with a linear execution history, preventing experimental state reuse. Appendix~\ref{app:ablation-definitions} details the retained and removed components.

% The ablation tests whether either form of exploration state suffices on its own. 
% On  20 Open set instances, \method completes 15 tasks (75.0\%), compared with 12 (60.0\%) for NoPool and 10 (50.0\%) for NoTree (Table~\ref{tab:ablation}). The three-way objective comparison gives \method 9 wins, versus 4 for NoPool and 2 for NoTree. Removing either component reduces both the number of successful tasks and the number of final-objective wins, supporting the complementary value of retaining alternative directions and reusable starting states.
As shown in Table~\ref{tab:ablation}, on 20 Open-set instances, \method successfully completes 15 tasks (75.0\%), compared with 12 (60.0\%) for NoPool and 10 (50.0\%) for NoTree. It also achieves the best final objective on 9 instances, versus 4 for NoPool and 2 for NoTree. The degradation of both ablations indicates that the idea pool and algorithm tree provide complementary benefits: the former preserves diverse research directions, while the latter enables effective reuse of compatible experimental states.

\begin{table}[!htbp]
\caption{Ablations on 20 shared instance identities.}
\label{tab:ablation}
\centering
\begingroup\fontsize{7}{8}\selectfont\setlength{\tabcolsep}{0.25pt}\renewcommand{\arraystretch}{2}
\begin{summarytabular}{16}{c*{3}{>{\columncolor{TableLavender}}c}*{3}{>{\columncolor{TableApricot}}c}*{3}{>{\columncolor{TableLavender}}c}*{3}{>{\columncolor{TableApricot}}c}*{3}{>{\columncolor{TableMint}}c}}\toprule
\multicolumn{1}{c}{\cellcolor{TableHeader}Method} & \multicolumn{3}{c}{\cellcolor{TableLavender}0--1 h} & \multicolumn{3}{c}{\cellcolor{TableApricot}1--3 h} & \multicolumn{3}{c}{\cellcolor{TableLavender}3--6 h} & \multicolumn{3}{c}{\cellcolor{TableApricot}6--12 h} & \multicolumn{3}{c}{\cellcolor{TableMint}Total} \\
\cmidrule(lr){2-4}\cmidrule(lr){5-7}\cmidrule(lr){8-10}\cmidrule(lr){11-13}\cmidrule(lr){14-16}
\multicolumn{1}{c}{\cellcolor{TableHeader}} & \multicolumn{1}{c}{\cellcolor{TableHeader}w/s} & \multicolumn{1}{c}{\cellcolor{TableHeader}time (s)} & \multicolumn{1}{c}{\cellcolor{TableHeader}token} & \multicolumn{1}{c}{\cellcolor{TableHeader}w/s} & \multicolumn{1}{c}{\cellcolor{TableHeader}time (s)} & \multicolumn{1}{c}{\cellcolor{TableHeader}token} & \multicolumn{1}{c}{\cellcolor{TableHeader}w/s} & \multicolumn{1}{c}{\cellcolor{TableHeader}time (s)} & \multicolumn{1}{c}{\cellcolor{TableHeader}token} & \multicolumn{1}{c}{\cellcolor{TableHeader}w/s} & \multicolumn{1}{c}{\cellcolor{TableHeader}time (s)} & \multicolumn{1}{c}{\cellcolor{TableHeader}token} & \multicolumn{1}{c}{\cellcolor{TableHeader}w/s} & \multicolumn{1}{c}{\cellcolor{TableHeader}time (s)} & \multicolumn{1}{c}{\cellcolor{TableHeader}token} \\ \midrule
\shortstack{AutoMIP} & 5/8 & 1,440 & 317,315 & 3/5 & 5,545 & 912,789 & 1/2 & 11,256 & 1,960,932 & 0/0 & 43,200 & 2,076,973 & 9/15 & 13,888 & 1,070,460 \\
NoPool & 1/5 & 2,019 & 233,353 & 2/4 & 5,643 & 450,314 & 1/2 & 11,460 & 1,168,343 & 0/1 & 42,334 & 2,874,266 & 4/12 & 21,829 & 1,558,655 \\
NoTree & 1/3 & 1,304 & 202,388 & 1/6 & 5,841 & 518,799 & 0/1 & 13,560 & 604,118 & 0/0 & 43,200 & 2,935,203 & 2/10 & 24,226 & 1,683,805 \\
\bottomrule\end{summarytabular}
\endgroup

\end{table}

% The joint design also achieves earlier completion with lower aggregate token usage. By three hours, the full skill completes 13 tasks, already exceeding either ablation's final twelve-hour count. Its Total token summary is 1,070,460, compared with 1,558,655 for NoPool and 1,683,805 for NoTree: reductions of 31.3\% and 36.4\%, respectively. The full skill therefore completes more tasks with a lower token aggregate than either component ablation.
The joint design also achieves earlier completion with substantially lower token consumption. Within the first three hours, \method completes 13 tasks, already exceeding the final 12-hour completion count of either ablation. It also uses the fewest tokens overall, with 1,070,460 total tokens, compared with 1,558,655 for NoPool and 1,683,805 for NoTree, representing reductions of 31.3\% and 36.4\%, respectively. 
% These results show that combining the Persistent Idea Pool and Algorithm Tree improves both search efficiency and resource utilization.

\FloatBarrier
\subsection{Results on MINLPLib}
\label{sec:minlplib-results}
% We further evaluate \method on MINLPLib to test whether its benefits extend to nonlinear optimization. \method completes 52 of 60 nonlinear tasks (86.7\%), compared with 45 for Codex (75.0\%), 40 for Loop, 29 for AutoEoH, and 31 for EvoX (Table~\ref{tab:minlplib}). Its successful set contains all 45 Codex successes plus seven additional instances. The final-objective comparison also favors \method, with 16 wins versus 11 for Codex, 9 for Loop, and 8 each for AutoEoH and EvoX. These results distinguish two gains: success on more nonlinear tasks and the lowest final objective on more instances in the five-method comparison.

We further evaluate \method on MINLPLib to assess whether its benefits extend to nonlinear optimization. As shown in Table~\ref{tab:minlplib}, \method successfully completes 52 of 60 instances (86.7\%), compared with 45 for Codex (75.0\%), 40 for Loop, 29 for AutoEoH, and 31 for EvoX. Its successful set includes all 45 instances solved by Codex, together with 7 additional nonlinear instances. In terms of final solution quality, \method also achieves the best objective on 16 instances, versus 11 for Codex, 9 for Loop, and 8 each for AutoEoH and EvoX. These results demonstrate that \method not only solves more nonlinear optimization problems within the budget, but also attains superior final objectives on a larger number of instances.
% By 10,800 s, \method reaches 50 completions, exceeding the final Codex count by five; the final count rises to 52 during continued exploration (Figure~\ref{fig:curve}(c)). On the 45 common Codex successes, \method is faster in 31 cases, with a median time ratio of 1.24. 
By 10,800 s, \method reaches 50 completed instances, already exceeding the final Codex count by 5; continued exploration further increases the total to 52 (Figure~\ref{fig:curve}(c)). On 45 instances successfully solved by both methods, \method finishes earlier in 31 cases, with an average speedup $\times$1.24.
% Average coverage over the full budget is 81.1\% versus 71.2\%. Across the three cohorts and four comparators, all twelve median paired time ratios exceed one.

\begin{table}[t]
\caption{MINLPLib, the supplied common 60-instance subset.}
\label{tab:minlplib}
\centering
\begingroup
\fontsize{7}{8}\selectfont
\setlength{\tabcolsep}{0.25pt}
\renewcommand{\arraystretch}{2}

\begin{summarytabular}{16}{c*{3}{>{\columncolor{TableLavender}}c}*{3}{>{\columncolor{TableApricot}}c}*{3}{>{\columncolor{TableLavender}}c}*{3}{>{\columncolor{TableApricot}}c}*{3}{>{\columncolor{TableMint}}c}}
\toprule

\multicolumn{1}{c}{\cellcolor{TableHeader}Method}
& \multicolumn{3}{c}{\cellcolor{TableLavender}0--1 h}
& \multicolumn{3}{c}{\cellcolor{TableApricot}1--3 h}
& \multicolumn{3}{c}{\cellcolor{TableLavender}3--6 h}
& \multicolumn{3}{c}{\cellcolor{TableApricot}6--12 h}
& \multicolumn{3}{c}{\cellcolor{TableMint}Total} \\

\cmidrule(lr){2-4}
\cmidrule(lr){5-7}
\cmidrule(lr){8-10}
\cmidrule(lr){11-13}
\cmidrule(lr){14-16}

\multicolumn{1}{c}{\cellcolor{TableHeader}}
& \multicolumn{1}{c}{\cellcolor{TableHeader}w/s}
& \multicolumn{1}{c}{\cellcolor{TableHeader}time (s)}
& \multicolumn{1}{c}{\cellcolor{TableHeader}token}

& \multicolumn{1}{c}{\cellcolor{TableHeader}w/s}
& \multicolumn{1}{c}{\cellcolor{TableHeader}time (s)}
& \multicolumn{1}{c}{\cellcolor{TableHeader}token}

& \multicolumn{1}{c}{\cellcolor{TableHeader}w/s}
& \multicolumn{1}{c}{\cellcolor{TableHeader}time (s)}
& \multicolumn{1}{c}{\cellcolor{TableHeader}token}

& \multicolumn{1}{c}{\cellcolor{TableHeader}w/s}
& \multicolumn{1}{c}{\cellcolor{TableHeader}time (s)}
& \multicolumn{1}{c}{\cellcolor{TableHeader}token}

& \multicolumn{1}{c}{\cellcolor{TableHeader}w/s}
& \multicolumn{1}{c}{\cellcolor{TableHeader}time (s)}
& \multicolumn{1}{c}{\cellcolor{TableHeader}token} \\

\midrule

\shortstack{AutoMIP}
& 13/40 & 1,076 & 203,122
& 3/10 & 5,895 & 623,950
& 0/1 & 19,140 & 1,051,147
& 0/1 & 41,089 & 3,376,242
& 16/52 & 8,182 & 763,362 \\

Codex
& 9/35 & 1,219 & 200,220
& 2/10 & 5,501 & 409,595
& 0/0 & - & -
& 0/0 & 43,200 & 2,927,059
& 11/45 & 12,428 & 916,826 \\

Loop
& 8/32 & 1,553 & 262,392
& 1/6 & 6,437 & 465,120
& 0/2 & 18,703 & 1,779,411
& 0/0 & 43,200 & 3,537,847
& 9/40 & 16,495 & 1,425,050 \\

AutoEoH
& 7/25 & 933 & 193,259
& 1/4 & 6,538 & 546,097
& 0/0 & - & -
& 0/0 & 43,200 & 3,513,110
& 8/29 & 23,145 & 1,932,038 \\

EvoX
& 8/31 & 1,111 & 218,111
& 0/0 & - & -
& 0/0 & - & -
& 0/0 & 43,200 & 3,241,573
& 8/31 & 21,454 & 1,679,451 \\

\bottomrule
\end{summarytabular}

\endgroup
\end{table}

% \FloatBarrier
\subsection{Research Agent Comparison}
\label{sec:generalization}
We evaluate \method as a reusable skill across different agents. On 20 instances, \method completes 15 tasks with Codex (GPT-5.5) and 12 with Claude Code (Claude Sonnet 5) (Table~\ref{tab:transfer}). Both configurations make early progress, with 9 and 7 completions in the first hour, rising to 14 and 10 by three hours. All 12 Claude Code successes are also completed under Codex, and the final-objective comparison yields 9 wins for Codex versus 6 for Claude Code. Together, these results demonstrate that AutoMIP supports successful optimization research under both Codex and Claude Code. We also attempted to run \method in DeepSeek Harness \citep{deepseek-harness2026}, Gemini CLI \citep{geminicli}, and DeerFlow \citep{deerflow}. In our tested configurations, these systems could not sustain the required twelve-hour runs, so these attempts did not yield completed evaluations and are not included in Table~\ref{tab:transfer}.

\begin{table}[t]
\caption{Cross-agent evaluation on 20 instances. Both agents use \method.}
\label{tab:transfer}
\centering
\begingroup\fontsize{7}{8}\selectfont\setlength{\tabcolsep}{0.25pt}\renewcommand{\arraystretch}{2}
\begin{summarytabular}{16}{c*{3}{>{\columncolor{TableLavender}}c}*{3}{>{\columncolor{TableApricot}}c}*{3}{>{\columncolor{TableLavender}}c}*{3}{>{\columncolor{TableApricot}}c}*{3}{>{\columncolor{TableMint}}c}}\toprule
\multicolumn{1}{c}{\cellcolor{TableHeader}Method} & \multicolumn{3}{c}{\cellcolor{TableLavender}0--1 h} & \multicolumn{3}{c}{\cellcolor{TableApricot}1--3 h} & \multicolumn{3}{c}{\cellcolor{TableLavender}3--6 h} & \multicolumn{3}{c}{\cellcolor{TableApricot}6--12 h} & \multicolumn{3}{c}{\cellcolor{TableMint}Total} \\
\cmidrule(lr){2-4}\cmidrule(lr){5-7}\cmidrule(lr){8-10}\cmidrule(lr){11-13}\cmidrule(lr){14-16}
\multicolumn{1}{c}{\cellcolor{TableHeader}} & \multicolumn{1}{c}{\cellcolor{TableHeader}w/s} & \multicolumn{1}{c}{\cellcolor{TableHeader}time (s)} & \multicolumn{1}{c}{\cellcolor{TableHeader}token} & \multicolumn{1}{c}{\cellcolor{TableHeader}w/s} & \multicolumn{1}{c}{\cellcolor{TableHeader}time (s)} & \multicolumn{1}{c}{\cellcolor{TableHeader}token} & \multicolumn{1}{c}{\cellcolor{TableHeader}w/s} & \multicolumn{1}{c}{\cellcolor{TableHeader}time (s)} & \multicolumn{1}{c}{\cellcolor{TableHeader}token} & \multicolumn{1}{c}{\cellcolor{TableHeader}w/s} & \multicolumn{1}{c}{\cellcolor{TableHeader}time (s)} & \multicolumn{1}{c}{\cellcolor{TableHeader}token} & \multicolumn{1}{c}{\cellcolor{TableHeader}w/s} & \multicolumn{1}{c}{\cellcolor{TableHeader}time (s)} & \multicolumn{1}{c}{\cellcolor{TableHeader}token} \\ \midrule
Codex & 6/9 & 1,385 & 312,966 & 2/5 & 5,089 & 944,082 & 1/1 & 11,085 & 1,688,996 & 0/0 & 43,200 & 4,641,312 & 9/15 & 13,250 & 1,621,633 \\
Claude Code & 2/7 & 1,819 & 305,213 & 2/3 & 6,326 & 1,182,345 & 2/2 & 13,433 & 1,679,399 & 0/0 & 43,200 & 4,680,348 & 6/12 & 20,209 & 2,324,256 \\
\bottomrule\end{summarytabular}
\endgroup

\end{table}

% \section{Limitations}
% The empirical scope of this study is optimization on the evaluated instance cohorts and within their specified wall-clock budgets. Applying \method to other research domains requires a task skill with executable experiments, reliable evaluation, and restorable artifacts. The interface experiment demonstrates operation under Codex and Claude Code; the comparisons against a base agent establish gains under Codex. The ablations assess the persistent pool and branching tree as complete components, without isolating individual switching thresholds or pool-generation rules. Adaptive control of these choices and evaluation in additional research domains are natural extensions.

% \section{Conclusion}
% \method organizes autoresearch through a persistent candidate pool and reusable experimental states. Its skill protocol coordinates refinement, switching, and recovery. The results show earlier completion on mixed-integer tasks, higher success rates on nonlinear tasks, and more final-objective wins on Open and MINLPLib. Together with the component ablations, these findings establish structured exploration state as a practical component of autoresearch.

\section{Conclusion}
% We present \method in this paper that organizes autoresearch through a persistent candidate pool and reusable experimental states. Its skill protocol coordinates refinement, switching, and recovery. The results show earlier completion on mixed-integer tasks, higher success rates on nonlinear tasks, and more final-objective wins on Open set of MIPLib and MINLPLib. Together with the component ablations, these findings establish structured exploration state as a practical component of autoresearch.

% The empirical scope of this study is limited to the evaluated optimization instance cohorts. Extending \method to other research domains requires task skills that support executable experiments, reliable evaluation, and restorable artifacts. The interface experiments demonstrate operation under both Codex and Claude Code, while comparisons with a base agent establish gains under Codex. The ablations evaluate the persistent pool and branching tree as complete components rather than isolating individual switching thresholds or pool-generation rules. Extending the framework to additional research domains and developing adaptive mechanisms for these control choices are natural directions for future work.

We present \method, a systematic autoresearch framework for MILP and MINLP optimization that couples a persistent idea pool with an Algorithm Tree to jointly manage diverse research directions and reusable experimental states. By selecting both what to explore and where to start, \method enables LLM agents to refine promising directions, switch to alternatives, and reuse accumulated artifacts within an end-to-end research loop. Experiments on MIPLIB and MINLPLib show that AutoMIP can consistently achieve earlier completion, higher success rates, and more final-objective wins among methods. 
Ablations further show that removing either component reduces task completion and final-objective performance, demonstrating their complementary roles in autoresearch.
The current empirical scope is focused on MIPLib and MINLPLib.  Extending \method to other research domains requires adjustments. 
% The framework operates with both Codex and Claude Code, while the comparative evaluation uses Codex. 
Future work includes developing auto-evolving task skills and algorithm search mechanisms for broader domains.

% \section*{AI Use Statement}

% We used large language models (LLMs) only to assist with literature discovery and to improve the grammar, clarity, and readability of the manuscript. Generative AI was not used to develop the core research ideas, algorithmic design, methodology, experimental framework, or conclusions of this work. All AI-assisted content was carefully reviewed and verified by the authors, who take full responsibility for the final manuscript, its claims, and the reported results.

% \clearpage
\bibliography{references}
\bibliographystyle{iclr2027_no_url}
\clearpage
\appendix
\section{Complete Skill Protocol}
\label{app:implementation}
This appendix specifies the \method skill. The skill controls experiment selection and memory; the primary task skill defines candidate construction, execution, and evaluation. Its four inputs are the snapshot interface $\mathcal{S}$, evaluator $E$, budget $B$, and stopping predicate $\tau$. The presentation follows the main text: the primary task skill is followed by PoolThink, algorithm-tree search, decision defaults, verification and recovery, and ablation definitions.

\subsection{Primary Task Skill: Opt}
\label{app:opt-protocol}
\begin{promptbox}{Opt: Task Objectives and Method Selection}

\textbf{Task and Input:} Understand the original MPS instance, determine its objective and constraints, seek a feasible or improved solution, and record attempts and results. Inputs are the MPS instance, task objective, and any known best or incumbent solution.

\textbf{Benchmark Context:} For MIPLIB instances, the official instance page may be consulted for status, reference objective, and problem family; it supplements rather than replaces direct MPS inspection.

\textbf{Open:} Find a feasible solution that strictly improves the current best-known or incumbent objective in the original optimization direction. An optimality proof is not required.

\textbf{Hard or Solved:} Reach the same known-best or optimal objective with shorter solve time, a faster recovery path, or a more reproducible strategy. An optimality proof is not required.

\textbf{Method Selection:} Choose methods based on MPS structure, variable and constraint patterns, objective, and known failure modes. The method space includes general-purpose solvers (e.g., Gurobi, HiGHS, SCIP, CPLEX, CBC, and OR-Tools), solver strategies, mathematical programming, heuristics, problem-specific algorithms, and hybrid methods. Do not assume that running a general-purpose solver on the full model is the only approach. State the solver's role, if used: baseline, verifier, subproblem solver, repair method, local-search tool, or primary solver. Reproducible ideas from prior work must be assessed against the current instance structure.

\textbf{Budget and Stopping:} Start timing when work begins. The maximum continuous budget is 43{,}200 seconds; a task may set a shorter budget. Stop when the objective is reached, the user requests termination, a safety boundary is triggered, or the budget expires. Record all time in seconds only.

\end{promptbox}

\begin{promptbox}{Opt: Workspace and MPS Analysis}

\textbf{Workspace:} Create a separate run directory for each instance under the project root's \texttt{optcodex-runs/}. Keep the original MPS read-only; record the instance, objective, run directory, and start time. Store inputs, code, models, logs, results, solutions, reports, and state in the workspace. Do not mix instances or scatter important artifacts elsewhere.

\textbf{MPS Inspection:} Inspect the MPS directly. Examine the objective and direction; row, column, nonzero, RHS, and bound counts; row and variable types; names and possible index sets; and matrix structure. Use these to infer possible problem families and structures, such as scheduling, routing, flow, assignment, coverage, inventory, temporal, block, or master--subproblem structure.

\textbf{Analysis Record:} Separate \texttt{Direct observations}, \texttt{Structure-based inferences}, and \texttt{Uncertain points}. Record implications for search keywords and subsequent idea generation. Use MIPLIB lookup only to check benchmark context, labels, known objectives, and problem family.

\end{promptbox}

\begin{promptbox}{Opt: Formulation, Verification, and Records}

\textbf{Implementation and Formulation:} Understand the instance and objective, select a method, and write the solving code. Derive the mathematical formulation from the implemented code, checking that variables, objective, constraints, indices, parameters, bounds, and domains agree. Prefer a semantic formulation; if business semantics cannot be recovered, use a structured formulation:

\begin{flushleft}
\(
\begin{array}{@{}ll}
\text{Sets: }& I,J,K. \qquad
\text{Parameters: } c_i,d_j,e_k,a_{ij},b_j,l_i,u_i,\ldots\\
\text{Variables: }&x_i\geq0,\quad y_j\in\{0,1\},\quad z_k\in\mathbb{Z}.\\
\text{Objective: }&
\min/\max\
\sum_{i\in I}c_ix_i+\sum_{j\in J}d_jy_j+\sum_{k\in K}e_kz_k.\\
\text{Constraints: }&
\sum_{i\in I}a_{ij}x_i\ \{=,\leq,\geq\}\ b_j\quad \forall j\in J,\\
&x_i\in[l_i,u_i]\quad\forall i\in I,\qquad
y_j\in\{0,1\}\quad\forall j\in J,\qquad
z_k\in\mathbb{Z}\quad\forall k\in K.
\end{array}
\)
\end{flushleft}
Use the row types and objective direction applicable to the instance.

\textbf{Consistency Check:} Verify variable correspondence, objective, constraints, index ranges, bounds, and types between code and formulation. Record the Epoch ID, Island ID, BEMT Node ID, problem type, sets, parameters, variables, MPS row/column interpretations, assumptions, and uncertainties.

\textbf{Independent Verification:} Export solution values, reload the original MPS, and check the candidate with Gurobi or an equivalent solver. Verify objective value, constraints, bounds, and integrality; save the solution and verification result, including pass status and maximum violation. Do not rely on solver logs alone or report an unverified solution as final.

\textbf{Outputs:} Preserve the problem summary, candidate directions, attempt outcomes, best verified result, failure reasons, and next steps. Keep scripts, logs, results, solution files, reports, and state organized in the instance workspace.

\textbf{Task Contract:} $\mathcal{C}=(\mathcal{S},E,B,\tau)$, where $\mathcal{S}$ is the executable candidate representation and required artifacts, $E$ is the evaluation and independent verification procedure, $B$ is the task-specific budget, and $\tau$ is the task-specific success and stopping condition.

\end{promptbox}

\subsection{Pool Construction and Lifecycle}
\label{app:pool}
\begin{promptbox}{PoolThink: Candidate Generation and Pool Lifecycle}

\textbf{Purpose:} The idea pool persistently stores complementary,
executable research directions for a single problem-solving run. It
separates candidate generation from the decision to execute a candidate.

\textbf{Initialization and Evidence Review:} At the start of a run,
PoolThink reviews the task structure and objective, baseline and best
verified results, constraints, code, execution logs, previous attempts,
and reusable artifacts. Literature review is optional when existing
evidence is sufficient; record the evidence sources used and why they
are sufficient.

\textbf{Candidate Generation and Screening:} Generate five to ten
complementary directions. Candidates should differ in their core
hypotheses or strategy families, such as neighborhood search,
structural decomposition, or model strengthening. Screen redundant
or infeasible proposals. Parameter, seed, or window changes within
the same hypothesis are refinements of that direction, not separate
candidates.

\textbf{Candidate Record:} Each pool record includes an ID, title,
status (\texttt{unused}, \texttt{selected}, \texttt{running},
\texttt{done}, or \texttt{rejected}), supporting evidence, hypothesis,
difference from attempted directions, starting requirements,
parent hint, and potential failure risk. The record also identifies
reusable artifacts required by the experiment, such as source code,
solver configurations, generated models, and intermediate files.

\textbf{Sampling and Use:} Let \(P_t\) be the pool at step \(t\), and
\[
U_t=\{q\in P_t:\operatorname{status}(q)=\texttt{unused}\}
\]
its unused candidates. When the Tree layer initiates an admissible
direction switch, it samples
\[
q_t\sim\operatorname{Uniform}(U_t).
\]
Registering the corresponding \(R\) branch consumes the candidate
identity for future switches. Later \(L\) attempts may continue
refining the selected direction. Execution and verification update
the candidate's status and accumulated evidence.

\textbf{Refresh:} Refresh the pool only after at least five distinct
candidates have been used or completed. A refresh repeats the state
review, failure analysis, evidence check, screening, and candidate
generation, while preserving the existing pool history. Pool history
persists within the run; a new problem instance begins with a new
task-specific review. If fewer than five candidates have been used
and no unused candidate is available, do not refresh: continue an
admissible \(L\) refinement, reject invalid candidates, or record why
progress is blocked.

\textbf{PoolThink Audit Record:} For each initialization or refresh,
record the current state and available artifacts (\texttt{State});
observed failures and risks (\texttt{Failures}); evidence sources
(\texttt{Evidence}); retained and rejected candidates with reasons
(\texttt{Screen}); generated directions (\texttt{Generate}); and their
IDs, statuses, and pool location (\texttt{Register}).

\end{promptbox}

\subsection{Branching Exploration Tree Protocol}
\label{app:tree-protocol}
\begin{promptbox}{Branching Exploration Tree: Search Control and Branching State}

\textbf{Tree Representation:} Let \(T_t=(V_t,D_t)\), where each node stores a research direction, observations, and reusable artifacts, and each edge records how a child experiment derives from its parent. The tree is multiway; \(L\) and \(R\) label edge semantics, not child positions.

\textbf{Edge Decisions:}
\begin{itemize}[leftmargin=*, itemsep=2pt, topsep=2pt]
  \item \textbf{\(L\), refinement:} Continue the same direction. Diagnostics, code checks, local repairs, model refinements, and parameter, seed, or window variants remain \(L\) when the core direction is unchanged. A failure alone does not justify \(R\).
  \item \textbf{\(R\), direction switch:} Change the core hypothesis, strategy family, or search path, or open a complementary branch from a historical state.
\end{itemize}

An \(R\) switch requires evidence: at least two related \(L\) attempts indicate the same stagnation or failure pattern; the core hypothesis is refuted by counterevidence, infeasibility, conflicting evidence, or resource constraints; no actionable \(L\) refinement remains; or an unused pool candidate is complementary to the observed failure. A single unexplained timeout or a vague impression of slow progress is insufficient.

\textbf{Candidate-Conditioned Parent Selection:} For \(L\), use the current node \(c_t\) as parent. Before \(R\), sample an unused pool candidate \(q_t\), then compare compatible parent states rather than defaulting to the current leaf. Let \(H_t\) contain the current node, its ancestors, and reusable frontier nodes. Let \(K_t(v,q_t)=1\) when node \(v\)'s artifacts satisfy candidate \(q_t\)'s starting requirements. Then
\[
C_t(q_t)=\{v\in H_t:K_t(v,q_t)=1\},\qquad
p_t=
\begin{cases}
c_t, & d_t=L,\\
\operatorname{SelectParent}(C_t(q_t),F_t,q_t), & d_t=R.
\end{cases}
\]
Parent selection considers candidate applicability, artifact dependencies, evidence value, unexplored complementary directions, and frontier priority. Each \(R\) decision records compatible parent alternatives, whether it backtracks or stays, the selected parent and rationale, and the consecutive-\(R\) check.

\textbf{Backtracking and Switch Chains:} Prefer a historical parent when the current leaf is a failed local repair, the path contains repeated failures, the candidate fits an ancestor better, or a compatible node is marked \texttt{backtrackable}. Stay when the candidate depends on the current node's artifacts or failure evidence and no better compatible parent exists. When the consecutive-\(R\) chain reaches two, prioritize a compatible ancestor, recent refresh node, or backtrackable frontier node. At length three, do not stay and add another \(R\) by default; allow this only when a strong artifact dependency leaves no compatible alternative. \(R\) siblings must use different pool candidates and have independent switch and parent rationales. Mark saturated parents in the frontier.

\textbf{Frontier Control:} After each node update, maintain \texttt{open}, \texttt{backtrackable}, \texttt{stalled}, and \texttt{closed} nodes, sibling summaries, priorities, and \texttt{right\_chain\_len}. The frontier must guide search: check \texttt{open} before \(L\), and compare ancestors, backtrackable nodes, and recent refresh nodes before \(R\). Set priorities according to evidence value, unexplored branches, failure complementarity, and artifact reuse.

\textbf{Registration:} Register the parent, edge label, branch rationale,
and selected pool candidate before execution. An \(R\) registration
consumes the selected candidate identity for future switches. Restore
the selected parent's artifacts and pass the registered experiment to
the verification layer. Record attempts that affect future decisions
in the tree, not only in logs or conversation.

\end{promptbox}

\subsection{Decision Rules and Defaults}
\label{app:policy}
\paragraph{Node granularity.}
Any attempt that informs a subsequent decision receives a node. Diagnostics, parameter sweeps, repairs, and failed experiments belong in the tree when they affect the next action. Each node records one coherent attempt at the level used by the primary skill. The relation between hypotheses determines its edge label.

\paragraph{Refinement ($L$).}
A trial remains an $L$ attempt when it continues the same direction, regardless of its size or outcome. Examples include modifying a parameter or seed, repairing code, refining a model, testing a different neighborhood size, and checking a counterexample. Evidence favoring $L$ includes a local improvement, a useful diagnosis, a plausible implementation issue, or a concrete next experiment that reduces uncertainty about the current hypothesis.

\paragraph{Switching ($R$).}
At least one of the following conditions justifies a direction switch:
\begin{enumerate}
\item At least two relevant $L$ nodes support the same stagnation or failure pattern.
\item A counterexample, infeasibility result, conflict, or resource constraint refutes the core hypothesis.
\item The current direction has no concrete next refinement.
\item An unused candidate directly complements an observed failure mode.
\end{enumerate}
An unexplained timeout or an attractive new idea alone does not satisfy this rule. The agent records the evidence supporting the switch before creating the node.

\paragraph{Parent selection.}
After sampling the unused candidate, the agent compares the current node, ancestors, recent refresh states, and \texttt{frontier.backtrackable}. A parent is compatible when its artifacts and assumptions support the proposed experiment. The \texttt{parent\_hint} guides this comparison. The required \texttt{backtrack\_check} records candidate parents and reasons, the decision to backtrack or stay, the selected parent, and the consecutive-$R$ ancestry check. Staying is appropriate when the selected experiment depends on artifacts specific to the current state. Backtracking is appropriate when earlier artifacts support the direction more directly or the current path repeats a failure pattern.

\paragraph{Repeated switches and siblings.}
After two consecutive $R$ edges, the next switch prioritizes an earlier compatible parent. After three, another $R$ from the current node requires a strong artifact dependency and no compatible alternative. The agent computes this check from parent ancestry. The CLI's \texttt{right\_chain\_len} records consecutive execution-order switches and is not a substitute for the ancestry check after backtracking. Multiple $R$ children may share a parent when they use distinct candidates and independent parent-selection rationales. Repeated branching from one parent triggers comparison with other backtrackable states; a saturated parent loses frontier priority.

\paragraph{Frontier maintenance.}
Open nodes have an actionable refinement. Backtrackable nodes offer reusable starting states. Stalled nodes lack current progress, and closed nodes are no longer eligible for expansion. Priorities reflect evidence value, unexplored alternatives, failure complementarity, and artifact reuse. After each attempt, the agent updates these categories and the sibling index before selecting the next action.

\subsection{Verification, State Update, and Recovery}
\label{app:verification-protocol}
\begin{promptbox}{Verification Layer: Execution, State Update, and Recovery}

\textbf{Inputs and Execution:} Receive the registered experiment, its
restored parent artifacts, the candidate interface \(\mathcal{S}\),
evaluator \(E\), remaining budget \(B\), and stopping predicate
\(\tau\). The primary task skill constructs and executes the candidate
from the restored state. The verification layer does not change the
registered parent or branch identity during execution.

\textbf{Evaluation and Independent Verification:} Evaluate feasibility,
objective value, runtime, and task-specific acceptance conditions using
\(E\). Export the candidate solution and verify it against the original
instance, including constraints, bounds, integrality, objective value,
pass status, and maximum violation. Do not treat solver logs alone as
verification or report an unverified result as final.

\textbf{State Update:} Save observations, validation evidence, generated
artifacts, failure reasons, and the next actionable step in the
registered node. Update the best verified result independently of the
active branch. Update the corresponding pool-candidate lifecycle,
frontier categories and priorities, current node, sibling summaries,
and timeline. A successful result may support another \(L\) refinement;
a diagnostic failure may motivate repair; repeated stagnation or
refutation may supply evidence for an \(R\) switch.

\textbf{Persistent State:}
\begin{itemize}[leftmargin=*, itemsep=2pt, topsep=2pt]
  \item \texttt{tree/nodes.jsonl}: minimal JSONL records with globally
        unique ID, parent ID (or \texttt{null} for root), side
        (\texttt{root}, \texttt{L}, or \texttt{R}), and concise action
        and rationale.
  \item \texttt{tree/nodes/<NodeID>/summary.md}: hypothesis, action,
        budget or stopping rule, observations, failure reason, next
        action, artifacts, and branch decision; include
        \texttt{backtrack\_check} for every \(R\) node.
  \item \texttt{tree/frontier.json}: current and prioritized frontier
        categories, sibling summaries, and \(R\)-chain length.
  \item \texttt{state/current.json}: current node, goal, last decision,
        next action, and update time.
  \item \texttt{pool/current\_pool.json}: candidate records and lifecycle
        status; \texttt{reports/timeline.md}: concise audit records for
        branches, resumes, refreshes, and closures.
\end{itemize}

\textbf{Validation and Recovery:} The state-writing script checks file
formats, parent existence, pool status, and recovery consistency; it
does not call an LLM or solver. Run the tree, frontier, pool, \(L/R\),
and recovery smoke test before actual solver runs. After interruption,
read \texttt{state/current.json}, \texttt{tree/nodes.jsonl}, the current
node and ancestors, relevant summaries, \texttt{tree/frontier.json}, and
\texttt{pool/current\_pool.json}, in that order. Restore the required
artifacts before selecting or executing the next attempt.

\textbf{Stopping and Final Report:} Stop when \(B\) is exhausted,
\(\tau\) is satisfied, or another task-specific stopping condition is
triggered. Report the best verified result, branch count, pool usage,
backtrack count, recurring failure patterns, elapsed time, and consumed
budget. Keep the timeline as an audit summary rather than duplicating
full experiment logs. Record time and budget in seconds.

\end{promptbox}

\subsection{Ablation Definitions}
\label{app:ablation-definitions}
We use two complementary ablations to evaluate the roles of persistent
candidate-pool management and branching state management. Both variants
use the primary task skill's candidate execution, evaluator, budget, and
stopping conditions.

\paragraph{NoPool.}
NoPool retains the multiway exploration tree, parent links, \(L/R\)
labels, frontier, node summaries, backtracking, and restoration of
artifacts from historical states. It removes PoolThink, the persistent
candidate pool, unused candidate records, candidate lifecycle tracking,
and random sampling from stored alternatives. When an \(R\) switch is
admissible, NoPool generates one complementary direction on demand from
the current evidence, including the current node, relevant ancestor
summaries, frontier, and observed failures. It does not retain
additional unused alternatives. The generated direction is recorded
directly in the \(R\) node; NoPool then uses the retained tree policy to
select a compatible parent and restore its artifacts. Subsequent
attempts that continue the selected direction are recorded as \(L\)
refinements. NoPool therefore evaluates tree-based branching and
historical-state reuse without persistent PoolThink-managed
alternatives.

\paragraph{NoTree.}
NoTree retains PoolThink, the persistent candidate pool, candidate
records and lifecycle states, and uniform selection from unused
directions. It removes tree nodes and parent links, \(L/R\) labels, the
frontier, sibling tracking, historical-parent selection, and
backtracking. Experiments are appended to one linear sequence. Each
experiment starts from the latest recorded state; local repairs may
continue the current strategy in that sequence, while a new strategy
is selected from the unused pool and appended as the next experiment.
Recovery resumes from the latest step, current state, and pool, without
reconstructing ancestry or selecting an earlier parent. In this
variant, the pool is refreshed after at least five candidates have
completed or been rejected.

\begin{table}[H]
\centering
\small
\begin{tabular*}{\linewidth}{@{\extracolsep{\fill}}lcc@{}}
\toprule
Mechanism & NoPool & NoTree \\
\midrule
PoolThink and persistent candidate pool & Removed & Retained \\
On-demand direction generation & Retained & Not used \\
Random selection from unused candidates & Removed & Retained \\
Multiway tree and parent links & Retained & Removed \\
\(L/R\) labels and frontier & Retained & Removed \\
Historical-state selection and backtracking & Retained & Removed \\
Experiment history & Branching tree & Linear sequence \\
\bottomrule
\end{tabular*}
\caption{Components retained and removed in the two complementary ablations.}
\label{tab:ablation-components}
\end{table}

\clearpage
\section{Task Prompts}
\label{app:task-prompts}
These reusable English templates specify the Open and Hard objectives and stopping rules. Replace \texttt{\{METHOD\}} with PoolTree, AutoEoH, Loop, or EvoX, and fill the instance, skill path, run directory, and objective. For plain Codex, omit the exploration-skill reading instruction and access permission; retain the same task, budget, verification, and isolation rules. These are standardized templates, not verbatim historical prompts.

\begin{promptbox}{Open Task: Incumbent Improvement}

\textbf{Skill and Instance:} Re-read \texttt{\{SKILL\_PATH\}} for
\texttt{\{METHOD\}} at the start and at regular intervals during the
run. Solve the Open instance \texttt{\{INSTANCE\_NAME\}}, available at
\texttt{\{INSTANCE\_URL\}}.

\textbf{Objective:} Find a feasible solution strictly better than the
starting best-known/incumbent objective
\texttt{\{REFERENCE\_OBJECTIVE\}}, respecting the original optimization
direction. If no feasible incumbent is known, find any feasible solution.
An optimality proof is not required. Check feasibility and the objective
against the original instance using the fixed evaluation tolerances.

\textbf{Budget and Stopping:} Work until this target is verified or
43,200 seconds have elapsed from task start. Do not stop early merely
because attempts fail. The budget includes preparation, skill reading,
coding, solving, and validation. Stop all run-owned computation at the
deadline, even if unsuccessful.

\textbf{Isolation:} This is an isolated comparison. Access only
\texttt{\{RUN\_DIR\}}, the supplied instance, and the assigned skill and
explicitly allowed dependencies. Do not inspect other experiment
directories, previous-run logs, other methods' results, or hidden
reference solutions.

\textbf{Outputs:} Save the best verified solution, objective, elapsed
time, and validation evidence in \texttt{\{RUN\_DIR\}}.

\end{promptbox}

\begin{promptbox}{Hard Task: Time to a Fixed Target}

\textbf{Skill and Instance:} Re-read \texttt{\{SKILL\_PATH\}} for
\texttt{\{METHOD\}} at the start and at regular intervals during the
run. Solve the Hard instance \texttt{\{INSTANCE\_NAME\}}, available at
\texttt{\{INSTANCE\_URL\}}.

\textbf{Objective:} Reach the supplied historical best-known objective
\texttt{\{TARGET\_OBJECTIVE\}} as quickly as possible. A feasible
solution that matches or improves this target in the original
optimization direction counts as success; a worse feasible solution
does not. An optimality proof is not required. Verify feasibility and
target attainment against the original instance using the fixed
evaluation tolerances.

\textbf{Budget and Stopping:} Work until the target is verified or
3,600 seconds have elapsed from task start. Do not stop early merely
because attempts fail. The budget includes preparation, skill reading,
coding, solving, and validation. Stop all run-owned computation at the
deadline, even if the target remains unmet.

\textbf{Isolation:} This is an isolated comparison. Access only
\texttt{\{RUN\_DIR\}}, the supplied instance, and the assigned skill and
explicitly allowed dependencies. Do not inspect other experiment
directories, previous-run logs, other methods' results, or hidden
reference solutions.

\textbf{Outputs:} Save the best verified solution, objective, time to
verified target attainment (if reached), elapsed time, and validation
evidence in \texttt{\{RUN\_DIR\}}.

\end{promptbox}

\clearpage
% Presentation-only fragment. Requires graphicx; keep the simulated-data caption.
% Vector asset: generated/simulated_instance_convergence.pdf; all trajectories are synthetic.
\section{Illustrative Instance-Level Convergence}
\label{app:simulated-convergence}

\begin{figure}[H]
  \centering
  \includegraphics[width=\linewidth]{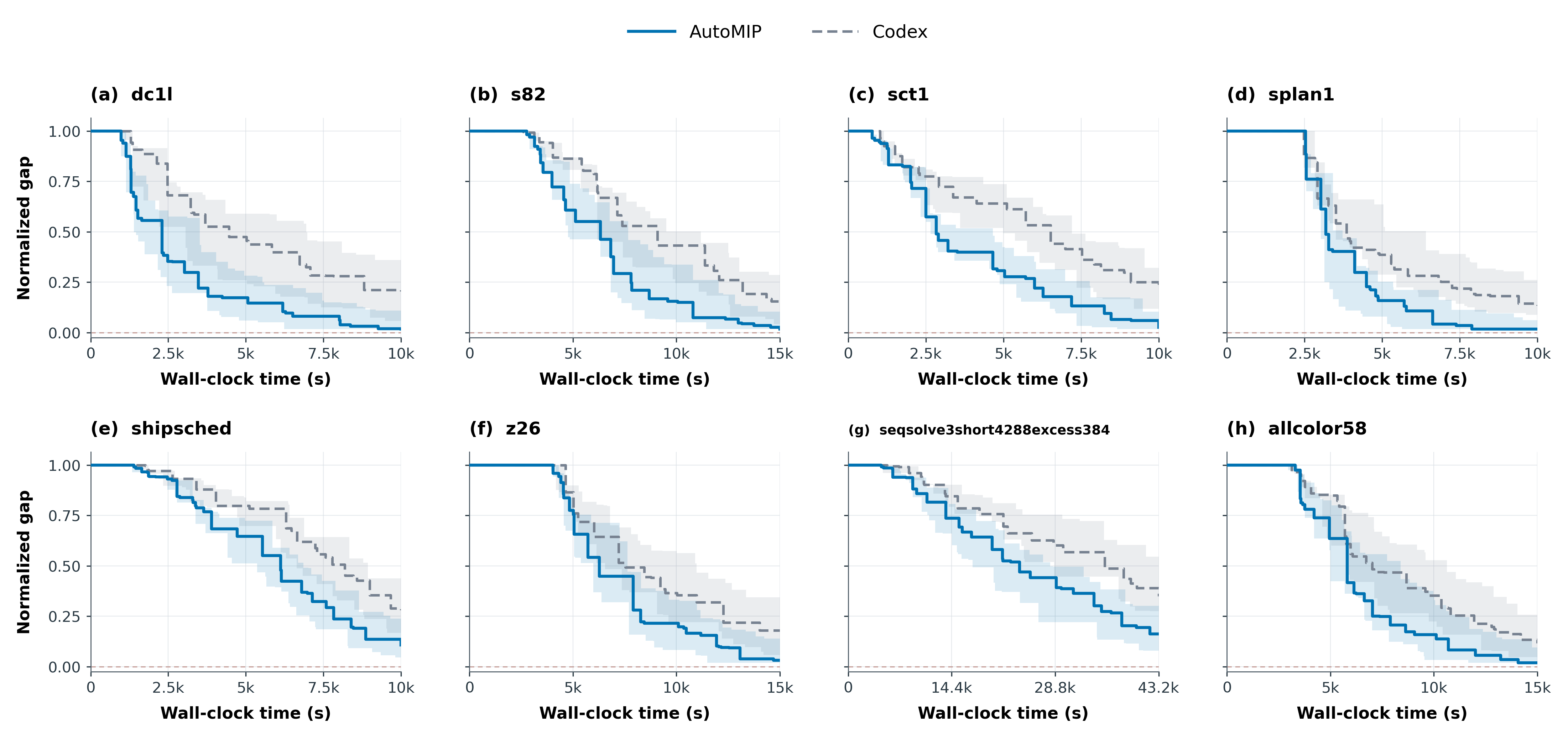}
  \caption{Instance-level convergence over ten trajectories per instance: five random seeds for each of the two methods. Lines show each method's pointwise median; shaded bands show its interquartile range (25th--75th percentiles) across the five seeds.}
  \label{fig:instance-convergence}
\end{figure}

Figure~\ref{fig:instance-convergence} summarizes ten trajectories per instance, comprising five random seeds for each of the two methods. At every time point, the curve reports the median of the five seed-specific trajectories, and the shaded band spans their 25th to 75th percentiles. This presentation makes the typical trajectory and variation across seeds visible throughout the search. In the displayed \texttt{dc1l}, \texttt{sct1}, and \texttt{splan1} trajectories, the \method curve enters the lower-gap region through pronounced early drops. The \texttt{s82}, \texttt{z26}, and \texttt{allcolor58} panels show successive improvements following an initial plateau, while \texttt{shipsched} exhibits a more gradual decline through the middle and later portions of the window. The longest panel, \texttt{seqsolve3short4288excess384}, extends to 43,200 s and retains positive terminal values for both curves. Its continuing descent emphasizes progress accumulated throughout a long search. Across these illustrations, the separation develops over multiple intermediate steps and remains visible at the end of the window.

These two regimes motivate the complementary decisions encoded by the \method skill. Same-direction refinement uses $L$ transitions to build on productive hypotheses and reusable artifacts. When accumulated evidence supports a change of direction, the persistent pool supplies an unused complementary candidate, and the tree identifies an artifact-compatible parent from which to proceed. Frontier updates preserve promising continuation and backtracking opportunities across these decisions. The control policy is therefore designed to combine sustained refinement with renewed exploration: continue a productive direction while maintaining concrete alternatives for the next switch.

\clearpage
\section{Per-Instance Open Results}
\label{app:open}
\begingroup\small
Tables~\ref{tab:instances-open-0}--\ref{tab:instances-open-1} list the 60 Open instances across all five methods. YES and NO preserve the source outcome labels. In all per-instance tables, a dash following \texttt{NO} means no successful result was recorded, so no corresponding time or token value is available. All times are in seconds. The main analysis counts successes within the 43,200 s horizon using these timestamps.
\par\endgroup
\begingroup
\setlength{\intextsep}{1pt}
\setlength{\abovecaptionskip}{3pt}
\begin{table}[H]
\centering
\caption{Per-instance Open results, panel 1.}
\label{tab:instances-open-0}
\fontsize{8.5}{9}\selectfont
\setlength{\tabcolsep}{2pt}
\renewcommand{\arraystretch}{0.95}
\begin{tabular*}{\linewidth}{@{\extracolsep{\fill}}l*{3}{crr}@{}}
\toprule
Instance & \multicolumn{3}{c}{AutoMIP} & \multicolumn{3}{c}{Codex} & \multicolumn{3}{c}{Loop} \\
\cmidrule(lr){2-4}\cmidrule(lr){5-7}\cmidrule(lr){8-10}
 & Result & Time (s) & Token & Result & Time (s) & Token & Result & Time (s) & Token \\ \midrule
allcolor58 & YES & 7,491 & 2,242,663 & YES & 11,061 & 3,272,799 & YES & 11,460 & 944,874 \\
cmflsp40-36-2-10 & YES & 1,707 & 389,130 & YES & 5,460 & 558,205 & YES & 32,772 & 4,151,482 \\
dc1l & YES & 1,713 & 457,746 & YES & 8,612 & 3,614,436 & NO & - & - \\
neos-1423785 & YES & 3,660 & 356,572 & YES & 43,200 & 2,819,091 & YES & 6,780 & 382,582 \\
nsr8k & YES & 2,467 & 670,627 & YES & 5,460 & 951,526 & YES & 5,580 & 104,652 \\
s82 & YES & 172 & 76,005 & YES & 13,500 & 6,038,343 & NO & - & - \\
supportcase39 & YES & 3,313 & 820,920 & YES & 30,540 & 3,904,230 & YES & 1,966 & 471,842 \\
supportcase41 & YES & 187 & 212,308 & YES & 977 & 278,223 & YES & 1,592 & 205,822 \\
triptim4 & NO & - & - & NO & - & - & NO & - & - \\
zeil & YES & 11,085 & 1,688,996 & YES & 5,209 & 887,698 & NO & - & - \\
z26 & YES & 6,972 & 1,045,459 & YES & 13,118 & 1,397,971 & NO & - & - \\
xmas10-2 & NO & - & - & NO & - & - & NO & - & - \\
uccase10 & YES & 1,182 & 506,774 & YES & 3,194 & 1,138,567 & YES & 2,058 & 99,476 \\
t1722 & NO & - & - & NO & - & - & NO & - & - \\
cvrpp-n16k8vrpi & NO & - & - & NO & - & - & NO & - & - \\
core2586-950 & NO & - & - & NO & - & - & NO & - & - \\
bley\_xs1noM & NO & - & - & NO & - & - & NO & - & - \\
bley\_xs1 & NO & - & - & NO & - & - & NO & - & - \\
a2864-99blp & NO & - & - & NO & - & - & NO & - & - \\
assign1-10-4 & NO & - & - & NO & - & - & NO & - & - \\
d20200 & NO & - & - & NO & - & - & NO & - & - \\
dano3mip & NO & - & - & NO & - & - & NO & - & - \\
neos-2629914-sudost & NO & - & - & NO & - & - & NO & - & - \\
neos-2974461-ibar & NO & - & - & NO & - & - & NO & - & - \\
neos-2978205-isar & NO & - & - & NO & - & - & NO & - & - \\
ns1905797 & NO & - & - & NO & - & - & NO & - & - \\
tokyometro & NO & - & - & NO & - & - & NO & - & - \\
tpl-tub-ws1617 & NO & - & - & NO & - & - & NO & - & - \\
supportcase35 & YES & 630 & 260,812 & YES & 1,316 & 283,873 & YES & 1,521 & 186,649 \\
supportcase38 & YES & 2,663 & 444,368 & YES & 3,840 & 519,658 & NO & - & - \\
supportcase30 & NO & - & - & NO & - & - & NO & - & - \\
supportcase31 & YES & 644 & 119,183 & YES & 1,029 & 167,466 & NO & - & - \\
supportcase22 & NO & - & - & NO & - & - & NO & - & - \\
t1717 & NO & - & - & NO & - & - & NO & - & - \\
xmas10 & NO & - & - & NO & - & - & NO & - & - \\
supportcase23 & YES & 506 & 323,676 & NO & - & - & NO & - & - \\
stockholm & NO & 32,109 & 1,187,297 & YES & 10,440 & 366,359 & YES & 10,020 & 462,696 \\
splan1 & YES & 772 & 285,522 & YES & 5,777 & 382,132 & YES & 11,596 & 1,333,815 \\
sorrell7 & NO & - & - & NO & - & - & NO & - & - \\
snp-10-052-052 & YES & 806 & 149,313 & YES & 1,138 & 191,625 & YES & 1,380 & 347,690 \\
snp-10-004-052 & YES & 787 & 255,717 & YES & 938 & 275,343 & YES & 2,150 & 678,178 \\
snp-06-004-052 & YES & 1,022 & 293,466 & YES & 4,703 & 915,995 & YES & 1,740 & 539,928 \\
snp-04-052-052 & YES & 550 & 171,129 & YES & 581 & 183,134 & YES & 1,440 & 187,958 \\
sing17 & YES & 201 & 231,496 & YES & 2,620 & 533,820 & YES & 11,940 & 2,778,668 \\
sing5 & YES & 2,119 & 578,361 & YES & 1,989 & 459,759 & NO & - & - \\
sing11 & YES & 11,427 & 2,232,868 & YES & 2,506 & 283,011 & YES & 3,858 & 282,907 \\
siena1 & YES & 754 & 319,769 & YES & 18,240 & 719,581 & YES & 701 & 110,555 \\
shs1042 & YES & 1,400 & 102,233 & NO & - & - & YES & 2,177 & 144,302 \\
shs1014 & YES & 2,053 & 333,472 & YES & 2,144 & 334,990 & YES & 3,030 & 802,149 \\
shipsched & YES & 5,940 & 278,176 & YES & 3,105 & 529,245 & YES & 2,100 & 110,348 \\
set3-16 & YES & 1,191 & 187,389 & YES & 4,740 & 503,622 & YES & 1,332 & 185,508 \\
set3-09 & YES & 996 & 310,668 & YES & 2,149 & 460,154 & YES & 1,731 & 303,236 \\
seqsolve3short4288excess384 & NO & - & - & NO & - & - & YES & 9,780 & 738,216 \\
seqsolve2short4288 & NO & - & - & NO & - & - & NO & - & - \\
seqsolve1 & NO & - & - & NO & - & - & NO & - & - \\
sct5 & YES & 1,220 & 185,773 & YES & 8,400 & 617,122 & YES & 3,407 & 486,106 \\
sct1 & YES & 3,660 & 641,077 & YES & 8,040 & 2,461,717 & YES & 6,720 & 1,922,889 \\
scpn2 & NO & - & - & NO & - & - & NO & - & - \\
scpm1 & NO & - & - & NO & - & - & NO & - & - \\
scpl4 & NO & - & - & NO & - & - & NO & - & - \\
\bottomrule
\end{tabular*}
\end{table}
\endgroup
\clearpage
\begingroup
\setlength{\intextsep}{5pt}
\begin{table}[H]
\centering
\caption{Per-instance Open results, panel 2.}
\label{tab:instances-open-1}
\fontsize{8.5}{9}\selectfont
\setlength{\tabcolsep}{2pt}
\renewcommand{\arraystretch}{1}
\begin{tabular*}{\linewidth}{@{\extracolsep{\fill}}l*{2}{crr}@{}}
\toprule
Instance & \multicolumn{3}{c}{AutoEoH} & \multicolumn{3}{c}{EvoX} \\
\cmidrule(lr){2-4}\cmidrule(lr){5-7}
 & Result & Time (s) & Token & Result & Time (s) & Token \\ \midrule
allcolor58 & YES & 8,064 & 1,589,550 & NO & - & - \\
cmflsp40-36-2-10 & YES & 3,399 & 532,571 & NO & - & - \\
dc1l & NO & - & - & YES & 13,380 & 839,628 \\
neos-1423785 & YES & 3,761 & 393,219 & YES & 1,940 & 322,286 \\
nsr8k & YES & 13,800 & 3,076,397 & NO & - & - \\
s82 & NO & - & - & YES & 5,760 & 469,866 \\
supportcase39 & YES & 5,531 & 593,913 & NO & - & - \\
supportcase41 & YES & 1,680 & 197,120 & NO & - & - \\
triptim4 & NO & - & - & YES & 755 & 131,559 \\
zeil & YES & 13,980 & 496,139 & NO & - & - \\
z26 & NO & - & - & NO & - & - \\
xmas10-2 & NO & - & - & NO & - & - \\
uccase10 & YES & 3,420 & 501,324 & YES & 518 & 124,042 \\
t1722 & NO & - & - & NO & - & - \\
cvrpp-n16k8vrpi & NO & - & - & NO & - & - \\
core2586-950 & NO & - & - & NO & - & - \\
bley\_xs1noM & NO & - & - & NO & - & - \\
bley\_xs1 & NO & - & - & NO & - & - \\
a2864-99blp & NO & - & - & NO & - & - \\
assign1-10-4 & NO & - & - & NO & - & - \\
d20200 & NO & - & - & NO & - & - \\
dano3mip & NO & - & - & NO & - & - \\
neos-2629914-sudost & NO & - & - & NO & - & - \\
neos-2974461-ibar & NO & - & - & NO & - & - \\
neos-2978205-isar & NO & - & - & NO & - & - \\
ns1905797 & NO & - & - & NO & - & - \\
tokyometro & NO & - & - & NO & - & - \\
tpl-tub-ws1617 & NO & - & - & NO & - & - \\
supportcase35 & YES & 5,940 & 530,271 & YES & 1,906 & 333,042 \\
supportcase38 & NO & - & - & YES & 14,430 & 873,008 \\
supportcase30 & NO & - & - & NO & - & - \\
supportcase31 & YES & 42,394 & 1,058,044 & YES & 5,460 & 1,011,311 \\
supportcase22 & NO & - & - & NO & - & - \\
t1717 & NO & - & - & NO & - & - \\
xmas10 & NO & - & - & NO & - & - \\
supportcase23 & NO & - & - & NO & - & - \\
stockholm & YES & 10,320 & 722,848 & NO & - & - \\
splan1 & YES & 8,280 & 1,486,570 & YES & 5,700 & 538,138 \\
sorrell7 & NO & - & - & NO & - & - \\
snp-10-052-052 & YES & 2,976 & 2,310,932 & YES & 650 & 177,211 \\
snp-10-004-052 & YES & 2,580 & 979,307 & YES & 9,960 & 675,733 \\
snp-06-004-052 & YES & 1,980 & 797,414 & NO & - & - \\
snp-04-052-052 & YES & 2,400 & 810,052 & NO & - & - \\
sing17 & YES & 1,209 & 136,141 & NO & - & - \\
sing5 & YES & 10,363 & 2,401,321 & YES & 32,864 & 445,089 \\
sing11 & YES & 17,820 & 5,616,856 & NO & - & - \\
siena1 & YES & 3,616 & 1,346,109 & YES & 8,280 & 907,220 \\
shs1042 & YES & 8,400 & 252,910 & NO & - & - \\
shs1014 & NO & - & - & YES & 2,042 & 752,436 \\
shipsched & YES & 938 & 139,528 & NO & - & - \\
set3-16 & YES & 950 & 141,118 & NO & - & - \\
set3-09 & YES & 1,980 & 285,647 & YES & 12,360 & 3,047,204 \\
seqsolve3short4288excess384 & YES & 11,100 & 1,037,866 & NO & - & - \\
seqsolve2short4288 & NO & - & - & NO & - & - \\
seqsolve1 & NO & - & - & NO & - & - \\
sct5 & YES & 2,374 & 445,084 & NO & - & - \\
sct1 & NO & - & - & NO & - & - \\
scpn2 & NO & - & - & NO & - & - \\
scpm1 & NO & - & - & NO & - & - \\
scpl4 & NO & - & - & NO & - & - \\
\bottomrule
\end{tabular*}
\end{table}
\endgroup

\clearpage
\section{Per-Instance Hard Results}
\label{app:hard}
\begingroup\small
Tables~\ref{tab:instances-hard-0}--\ref{tab:instances-hard-1} list the 60 fixed-target tasks. Successes are counted within 3,600 s. Available token fields for unsuccessful runs remain visible. Table~\ref{tab:hard} retains the supplied aggregate time and token columns.
\par\endgroup
\begingroup
\setlength{\intextsep}{1pt}
\setlength{\abovecaptionskip}{3pt}
\begin{table}[H]
\centering
\caption{Per-instance Hard results, panel 1.}
\label{tab:instances-hard-0}
\fontsize{8.5}{9}\selectfont
\setlength{\tabcolsep}{2pt}
\renewcommand{\arraystretch}{0.95}
\begin{tabular*}{\linewidth}{@{\extracolsep{\fill}}l*{3}{crr}@{}}
\toprule
Instance & \multicolumn{3}{c}{AutoMIP} & \multicolumn{3}{c}{Codex} & \multicolumn{3}{c}{Loop} \\
\cmidrule(lr){2-4}\cmidrule(lr){5-7}\cmidrule(lr){8-10}
 & Result & Time (s) & Token & Result & Time (s) & Token & Result & Time (s) & Token \\ \midrule
2club200v15p5scn & YES & 1,048 & 118,047 & YES & 3,159 & 178,653 & YES & 720 & 98,788 \\
8div-n59k10 & NO & - & - & NO & - & - & NO & - & 187,464 \\
8div-n59k11 & NO & - & - & NO & - & 255,446 & NO & - & 126,442 \\
8div-n59k12 & NO & - & - & NO & - & 217,464 & NO & - & - \\
academictimetablebig & YES & 1,297 & 111,128 & YES & 3,113 & 160,989 & YES & 704 & 77,285 \\
adult-max5features & YES & 694 & 154,875 & YES & 310 & 831,951 & YES & 1,195 & 135,602 \\
adult-regularized & YES & 334 & 67,349 & YES & 496 & 70,269 & YES & 2,769 & 183,832 \\
allcolor10 & NO & - & - & YES & 962 & 157,564 & NO & - & - \\
amaze22012-07-04i & NO & - & - & NO & - & 117,032 & NO & - & 177,449 \\
atm20-100 & NO & - & - & NO & - & - & NO & - & 216,207 \\
bab1 & YES & 652 & 169,845 & YES & 347 & 94,893 & YES & 309 & 58,707 \\
bab2 & YES & 1,586 & 204,964 & YES & 2,216 & 170,933 & NO & - & 200,963 \\
bg512142 & NO & - & - & NO & - & 125,794 & NO & - & 135,260 \\
bppc6-02 & YES & 503 & 98,525 & YES & 245 & 51,425 & YES & 596 & 81,524 \\
bts4-cta & NO & - & - & NO & - & 72,432 & NO & - & - \\
comp16-3idx & NO & - & - & NO & - & - & NO & - & - \\
dc1c & YES & 1,347 & 125,170 & YES & 3,323 & 142,528 & NO & - & - \\
dg012142 & NO & - & - & NO & - & - & NO & - & 127,437 \\
dolom1 & YES & 313 & 91,382 & YES & 3,086 & 137,235 & YES & 591 & 104,627 \\
ds & NO & - & - & NO & - & 77,736 & NO & - & 195,660 \\
dws008-03 & YES & 2,088 & 261,083 & YES & 1,315 & 199,248 & NO & - & - \\
elitserienhandball13i & NO & - & - & NO & - & - & NO & - & 146,518 \\
fhnw-sq2 & YES & 439 & 167,200 & YES & 1,454 & 168,349 & YES & 1,144 & 162,127 \\
fhnw-sq3 & YES & 318 & 123,893 & YES & 952 & 121,517 & YES & 854 & 129,336 \\
ger50-17-ptp-pop-6t & NO & - & 231,594 & NO & - & 295,511 & NO & - & 246,896 \\
ger50\_17\_trans & NO & - & - & NO & - & 163,197 & YES & 1,657 & 124,697 \\
germany50-UUM & NO & - & - & NO & - & 216,723 & NO & - & - \\
gfd-schedulen180f7d50m30k18-16i & NO & - & 189,249 & NO & - & - & NO & - & 264,870 \\
gmut-75-50 & NO & - & 115,922 & NO & - & - & NO & - & - \\
graph20-80-1rand & YES & 1,119 & 72,377 & YES & 316 & 84,011 & YES & 604 & 61,694 \\
hgms-det & YES & 341 & 98,916 & YES & 765 & 91,557 & YES & 889 & 81,139 \\
icir97\_potential & YES & 284 & 75,201 & YES & 217 & 45,944 & YES & 871 & 71,701 \\
in & NO & - & - & NO & - & - & NO & - & 249,078 \\
ivu52 & YES & 3,173 & 269,220 & NO & - & - & NO & - & - \\
kosova1 & NO & - & 126,072 & NO & - & - & NO & - & - \\
l2p1i & NO & - & - & NO & - & - & NO & - & - \\
markshare2 & NO & - & - & NO & - & - & NO & - & - \\
milo-v13-4-3d-3-0 & YES & 2,040 & 242,757 & YES & 1,837 & 214,976 & NO & - & - \\
mkc & YES & 321 & 78,998 & YES & 893 & 86,753 & YES & 1,252 & 161,679 \\
moj-mining & YES & 1,596 & 164,379 & YES & 1,943 & 192,288 & NO & - & 217,225 \\
neos-1140050 & NO & - & - & NO & - & 113,724 & NO & - & 154,115 \\
neos-3068746-nene & YES & 352 & 71,324 & YES & 1,761 & 147,347 & YES & 1,311 & 155,228 \\
neos-3209462-rhin & YES & 337 & 83,652 & YES & 2,499 & 100,270 & YES & 574 & 76,379 \\
neos-3211096-shag & NO & - & - & NO & - & 234,999 & NO & - & 439,013 \\
neos-3214367-sovi & NO & - & 291,348 & NO & - & - & NO & - & 244,482 \\
neos-3237086-abava & YES & 700 & 136,643 & NO & - & - & NO & - & - \\
neos-3322547-alsek & YES & 924 & 109,681 & YES & 1,364 & 132,797 & NO & - & 218,089 \\
neos-3352863-ancoa & NO & - & - & YES & 1,462 & 119,891 & NO & - & 161,669 \\
neos-3372571-onahau & YES & 626 & 83,080 & YES & 917 & 86,238 & YES & 2,651 & 186,730 \\
neos-3402454-bohle & NO & - & - & NO & - & - & NO & - & - \\
neos-3740487-motru & YES & 480 & 127,117 & YES & 2,423 & 180,750 & NO & - & - \\
neos-4295773-pissa & NO & - & - & NO & - & 126,952 & NO & - & 279,831 \\
neos-4321076-ruwer & NO & - & - & NO & - & 102,531 & YES & 1,941 & 196,930 \\
neos-4335793-snake & YES & 1,188 & 117,904 & YES & 2,373 & 100,930 & NO & - & - \\
neos-4409277-trave & YES & 1,728 & 167,212 & NO & - & - & YES & 2,068 & 208,452 \\
neos-4647027-thurso & YES & 386 & 81,424 & YES & 2,133 & 125,928 & YES & 1,234 & 101,056 \\
neos-4647030-tutaki & YES & 595 & 104,757 & YES & 967 & 158,366 & YES & 2,651 & 265,269 \\
neos-4954274-beardy & NO & - & 225,685 & NO & - & - & NO & - & 208,088 \\
neos-5196530-nuhaka & YES & 589 & 87,231 & YES & 2,993 & 168,960 & NO & - & - \\
neos-5223573-tarwin & NO & - & - & NO & - & - & YES & 2,976 & 309,145 \\
\bottomrule
\end{tabular*}
\end{table}
\endgroup
\clearpage
\begingroup
\setlength{\intextsep}{5pt}
\begin{table}[H]
\centering
\caption{Per-instance Hard results, panel 2.}
\label{tab:instances-hard-1}
\fontsize{8.5}{9}\selectfont
\setlength{\tabcolsep}{2pt}
\renewcommand{\arraystretch}{1}
\begin{tabular*}{\linewidth}{@{\extracolsep{\fill}}l*{2}{crr}@{}}
\toprule
Instance & \multicolumn{3}{c}{AutoEoH} & \multicolumn{3}{c}{EvoX} \\
\cmidrule(lr){2-4}\cmidrule(lr){5-7}
 & Result & Time (s) & Token & Result & Time (s) & Token \\ \midrule
2club200v15p5scn & YES & 1,099 & 189,724 & YES & 1,660 & 155,633 \\
8div-n59k10 & NO & - & 259,168 & NO & - & - \\
8div-n59k11 & NO & - & - & NO & - & 114,174 \\
8div-n59k12 & NO & - & - & NO & - & 153,108 \\
academictimetablebig & NO & - & 100,525 & NO & - & - \\
adult-max5features & YES & 892 & 104,780 & NO & - & 202,821 \\
adult-regularized & YES & 531 & 106,018 & YES & 586 & 122,327 \\
allcolor10 & NO & - & - & NO & - & - \\
amaze22012-07-04i & NO & - & 60,064 & NO & - & 295,929 \\
atm20-100 & NO & - & - & NO & - & 220,566 \\
bab1 & YES & 713 & 79,137 & YES & 568 & 109,137 \\
bab2 & NO & - & - & NO & - & 141,719 \\
bg512142 & NO & - & 193,816 & NO & - & 263,482 \\
bppc6-02 & YES & 424 & 89,289 & YES & 886 & 90,659 \\
bts4-cta & YES & 1,148 & 121,413 & YES & 567 & 121,862 \\
comp16-3idx & NO & - & - & NO & - & - \\
dc1c & NO & - & - & NO & - & - \\
dg012142 & NO & - & 153,341 & NO & - & 317,165 \\
dolom1 & NO & - & - & NO & - & 184,683 \\
ds & NO & - & 175,372 & NO & - & - \\
dws008-03 & YES & 1,211 & 167,231 & NO & - & - \\
elitserienhandball13i & NO & - & - & NO & - & - \\
fhnw-sq2 & YES & 2,302 & 273,012 & NO & - & - \\
fhnw-sq3 & NO & - & 107,761 & YES & 1,133 & 119,149 \\
ger50-17-ptp-pop-6t & NO & - & - & NO & - & - \\
ger50\_17\_trans & NO & - & 166,150 & YES & 1,351 & 190,891 \\
germany50-UUM & NO & - & - & NO & - & - \\
gfd-schedulen180f7d50m30k18-16i & NO & - & 127,457 & NO & - & - \\
gmut-75-50 & NO & - & - & NO & - & 242,651 \\
graph20-80-1rand & NO & - & 78,778 & YES & 692 & 98,261 \\
hgms-det & YES & 851 & 82,331 & NO & - & 208,646 \\
icir97\_potential & YES & 2,562 & 124,174 & NO & - & 272,504 \\
in & NO & - & - & YES & 2,589 & 221,296 \\
ivu52 & NO & - & - & NO & - & 156,888 \\
kosova1 & NO & - & 153,148 & NO & - & - \\
l2p1i & NO & - & - & YES & 2,816 & 305,698 \\
markshare2 & NO & - & 165,088 & NO & - & - \\
milo-v13-4-3d-3-0 & NO & - & - & NO & - & 220,694 \\
mkc & YES & 1,299 & 171,274 & NO & - & - \\
moj-mining & YES & 3,223 & 217,139 & NO & - & - \\
neos-1140050 & NO & - & - & NO & - & - \\
neos-3068746-nene & YES & 832 & 83,419 & YES & 1,598 & 195,704 \\
neos-3209462-rhin & NO & - & 174,006 & NO & - & 111,020 \\
neos-3211096-shag & NO & - & 138,113 & NO & - & 165,861 \\
neos-3214367-sovi & NO & - & - & NO & - & - \\
neos-3237086-abava & NO & - & - & NO & - & 129,706 \\
neos-3322547-alsek & YES & 1,729 & 141,824 & YES & 883 & 115,881 \\
neos-3352863-ancoa & YES & 2,058 & 117,127 & NO & - & - \\
neos-3372571-onahau & NO & - & - & NO & - & 143,493 \\
neos-3402454-bohle & NO & - & - & NO & - & 143,241 \\
neos-3740487-motru & NO & - & - & NO & - & 152,293 \\
neos-4295773-pissa & NO & - & 211,866 & NO & - & 522,515 \\
neos-4321076-ruwer & NO & - & - & NO & - & - \\
neos-4335793-snake & NO & - & - & NO & - & - \\
neos-4409277-trave & YES & 2,170 & 131,023 & YES & 1,574 & 271,273 \\
neos-4647027-thurso & YES & 1,645 & 91,438 & NO & - & - \\
neos-4647030-tutaki & YES & 1,962 & 117,914 & YES & 1,137 & 94,741 \\
neos-4954274-beardy & NO & - & - & NO & - & 277,155 \\
neos-5196530-nuhaka & YES & 2,286 & 131,077 & NO & - & - \\
neos-5223573-tarwin & NO & - & - & YES & 1,035 & 91,997 \\
\bottomrule
\end{tabular*}
\end{table}
\endgroup

\clearpage
\section{Per-Instance Ablation Results}
\label{app:ablation}
\begingroup\small
Table~\ref{tab:instances-ablation_full-0} gives the 20 shared instance identities for the full skill, NoPool, and NoTree. The full-skill column is the corresponding subset of Open observations. All three columns use the 43,200 s analysis horizon.
\par\endgroup
\begingroup
\setlength{\intextsep}{5pt}
\begin{table}[H]
\centering
\caption{Per-instance Ablation results.}
\label{tab:instances-ablation_full-0}
\fontsize{9}{11}\selectfont
\setlength{\tabcolsep}{1.5pt}
\renewcommand{\arraystretch}{1}
\begin{tabular*}{\linewidth}{@{\extracolsep{\fill}}l*{3}{crr}@{}}
\toprule
Instance & \multicolumn{3}{c}{AutoMIP} & \multicolumn{3}{c}{NoPool} & \multicolumn{3}{c}{NoTree} \\
\cmidrule(lr){2-4}\cmidrule(lr){5-7}\cmidrule(lr){8-10}
 & Result & Time (s) & Token & Result & Time (s) & Token & Result & Time (s) & Token \\ \midrule
allcolor58 & YES & 7,491 & 2,242,663 & YES & 10,920 & 1,091,491 & NO & - & - \\
dc1l & YES & 1,713 & 457,746 & NO & - & - & NO & - & - \\
neos-1423785 & YES & 3,660 & 356,572 & NO & - & - & YES & 13,560 & 604,118 \\
s82 & YES & 172 & 76,005 & YES & 12,000 & 1,245,195 & YES & 1,531 & 172,552 \\
triptim4 & NO & - & - & NO & - & - & NO & - & - \\
zeil & YES & 11,085 & 1,688,996 & YES & 3,380 & 325,760 & YES & 5,160 & 374,765 \\
z26 & YES & 6,972 & 1,045,459 & NO & - & - & YES & 3,876 & 519,591 \\
cvrpp-n16k8vrpi & NO & - & - & NO & - & - & NO & - & - \\
core2586-950 & NO & - & - & NO & - & - & NO & - & - \\
neos-2978205-isar & NO & - & - & NO & - & - & NO & - & - \\
supportcase35 & YES & 630 & 260,812 & YES & 685 & 115,381 & YES & 5,100 & 507,597 \\
supportcase38 & YES & 2,663 & 444,368 & YES & 4,980 & 585,617 & YES & 3,930 & 498,680 \\
splan1 & YES & 772 & 285,522 & YES & 766 & 143,269 & YES & 1,260 & 251,956 \\
sing5 & YES & 2,119 & 578,361 & YES & 9,960 & 446,297 & NO & - & - \\
sing11 & YES & 11,427 & 2,232,868 & YES & 3,732 & 441,758 & YES & 6,420 & 352,677 \\
shs1042 & YES & 1,400 & 102,233 & YES & 3,180 & 355,955 & NO & - & - \\
shs1014 & YES & 2,053 & 333,472 & YES & 3,900 & 327,582 & YES & 1,121 & 182,655 \\
shipsched & YES & 5,940 & 278,176 & YES & 2,083 & 226,400 & YES & 10,560 & 859,482 \\
seqsolve3short4288excess384 & NO & - & - & NO & - & - & NO & - & - \\
sct1 & YES & 3,660 & 641,077 & YES & 35,403 & 908,939 & NO & - & - \\
\bottomrule
\end{tabular*}
\end{table}
\endgroup

\clearpage
\section{Per-Instance Interface Results}
\label{app:transfer}
\begingroup\small
Table~\ref{tab:instances-transfer-0} records \method under Codex and Claude Code. Both columns use the skill and a 43,200 s horizon. These interface observations are kept separate from the Open and ablation records with the same instance identities.
\par\endgroup
\begingroup
\setlength{\intextsep}{5pt}
\begin{table}[H]
\centering
\caption{Per-instance CLI-transfer results.}
\label{tab:instances-transfer-0}
\fontsize{9}{11}\selectfont
\setlength{\tabcolsep}{1.5pt}
\renewcommand{\arraystretch}{1}
\begin{tabular*}{\linewidth}{@{\extracolsep{\fill}}l*{2}{crr}@{}}
\toprule
Instance & \multicolumn{3}{c}{Codex} & \multicolumn{3}{c}{Claude Code} \\
\cmidrule(lr){2-4}\cmidrule(lr){5-7}
 & Result & Time (s) & Token & Result & Time (s) & Token \\ \midrule
allcolor58 & YES & 7,491 & 2,242,663 & YES & 9,120 & 2,942,663 \\
dc1l & YES & 1,713 & 457,746 & NO & - & - \\
s82 & YES & 172 & 76,005 & YES & 15,420 & 2,743,724 \\
triptim4 & NO & - & - & NO & - & - \\
zeil & YES & 11,085 & 1,688,996 & YES & 3,078 & 412,299 \\
z26 & YES & 6,972 & 1,045,459 & NO & - & - \\
neos-1423785 & YES & 3,660 & 356,572 & YES & 1,160 & 221,741 \\
cvrpp-n16k8vrpi & NO & - & - & NO & - & - \\
core2586-950 & NO & - & - & NO & - & - \\
neos-2978205-isar & NO & - & - & NO & - & - \\
supportcase35 & YES & 630 & 260,812 & YES & 693 & 167,785 \\
supportcase38 & YES & 2,663 & 444,368 & YES & 1,917 & 398,586 \\
shipsched & YES & 940 & 278,176 & YES & 1,629 & 367,845 \\
splan1 & YES & 772 & 285,522 & YES & 3,043 & 434,652 \\
sing5 & YES & 2,119 & 578,361 & NO & - & - \\
sing11 & YES & 3,660 & 434,638 & YES & 5,254 & 316,520 \\
shs1042 & YES & 1,400 & 102,233 & YES & 4,605 & 287,853 \\
shs1014 & YES & 2,053 & 333,472 & YES & 1,214 & 133,584 \\
seqsolve3short4288excess384 & NO & - & - & NO & - & - \\
sct1 & YES & 3,660 & 641,077 & YES & 11,446 & 615,073 \\
\bottomrule
\end{tabular*}
\end{table}
\endgroup

\clearpage
\section{Per-Instance MINLPLib Results}
\label{app:minlplib}
\begingroup\small
Tables~\ref{tab:instances-minlplib-0}--\ref{tab:instances-minlplib-1} list the retained common 60-instance nonlinear cohort across all five methods. The outcome, time, and token fields support Table~\ref{tab:minlplib}, Figure~\ref{fig:curve}(c), and the paired and token analyses.
\par\endgroup
\begingroup
\setlength{\intextsep}{5pt}
\begin{table}[H]
\centering
\caption{Per-instance MINLPLib results, panel 1.}
\label{tab:instances-minlplib-0}
\fontsize{8.5}{9}\selectfont
\setlength{\tabcolsep}{2pt}
\renewcommand{\arraystretch}{1}
\begin{tabular*}{\linewidth}{@{\extracolsep{\fill}}l*{3}{crr}@{}}
\toprule
Instance & \multicolumn{3}{c}{AutoMIP} & \multicolumn{3}{c}{Codex} & \multicolumn{3}{c}{Loop} \\
\cmidrule(lr){2-4}\cmidrule(lr){5-7}\cmidrule(lr){8-10}
 & Result & Time (s) & Token & Result & Time (s) & Token & Result & Time (s) & Token \\ \midrule
elec100 & NO & - & - & NO & - & - & NO & - & - \\
gabriel09 & YES & 1,012 & 108,534 & YES & 4,154 & 242,505 & YES & 2,294 & 156,847 \\
etamac & YES & 1,226 & 358,589 & YES & 611 & 200,631 & YES & 699 & 205,836 \\
eq6\_1 & YES & 660 & 122,384 & YES & 927 & 121,620 & YES & 754 & 316,728 \\
emfl050 & YES & 1,219 & 215,608 & YES & 3,230 & 451,415 & YES & 2,678 & 423,937 \\
csched2 & YES & 19,140 & 1,051,147 & YES & 5,341 & 234,122 & YES & 1,205 & 221,530 \\
eg\_disc2 & YES & 6,900 & 317,886 & NO & - & - & NO & - & - \\
edgecross24-115 & NO & - & - & NO & - & - & NO & - & - \\
dtoc5 & YES & 461 & 176,052 & YES & 433 & 108,331 & YES & 1,155 & 311,534 \\
deb10 & YES & 654 & 182,053 & YES & 734 & 269,264 & YES & 869 & 225,698 \\
pooling\_ct3 & YES & 5,340 & 357,397 & YES & 6,780 & 642,700 & YES & 3,968 & 268,093 \\
crudeoil\_li01 & YES & 3,251 & 283,999 & YES & 3,472 & 292,765 & YES & 17,820 & 751,336 \\
contvar & YES & 758 & 330,630 & YES & 2,569 & 170,096 & YES & 7,929 & 546,068 \\
color\_lab6b\_4x20 & YES & 4,012 & 268,186 & NO & - & - & YES & 5,085 & 270,194 \\
color\_lab3\_4x0 & YES & 1,663 & 267,996 & NO & - & - & NO & - & - \\
color\_lab2\_4x0 & YES & 4,380 & 313,957 & YES & 4,524 & 255,851 & NO & - & - \\
chp\_shorttermplan2d & YES & 3,180 & 455,533 & YES & 3,071 & 358,004 & NO & - & - \\
chp\_shorttermplan2b & YES & 1,260 & 303,600 & YES & 5,440 & 1,077,661 & YES & 2,853 & 271,882 \\
chimera\_mgw-c16-2031-01 & NO & - & - & NO & - & - & NO & - & - \\
chimera\_mgw-c8-507-onc8-01 & NO & - & - & NO & - & - & NO & - & - \\
chimera\_rfr-01 & NO & - & - & NO & - & - & NO & - & - \\
chimera\_mgw-c16-2031-02 & NO & - & - & NO & - & - & NO & - & - \\
chimera\_lga-02 & NO & - & - & NO & - & - & NO & - & - \\
chain50 & YES & 674 & 160,312 & YES & 1,799 & 309,002 & YES & 968 & 89,302 \\
chain400 & YES & 473 & 160,844 & YES & 1,479 & 233,991 & YES & 1,614 & 158,087 \\
chain200 & YES & 584 & 175,414 & YES & 1,413 & 161,270 & YES & 834 & 133,212 \\
chain100 & YES & 705 & 171,110 & YES & 1,030 & 175,714 & YES & 19,586 & 2,807,486 \\
cesam2log & YES & 1,340 & 234,194 & YES & 857 & 207,052 & YES & 440 & 101,429 \\
cesam2cent & YES & 986 & 250,919 & YES & 4,660 & 283,297 & YES & 1,687 & 329,771 \\
catmix100 & YES & 951 & 144,276 & YES & 768 & 141,165 & NO & - & - \\
case\_1scv2 & YES & 665 & 237,478 & YES & 491 & 202,184 & YES & 3,085 & 315,687 \\
blendgap & NO & - & - & NO & - & - & NO & - & - \\
casctanks & YES & 4,600 & 665,727 & YES & 1,405 & 255,220 & YES & 2,543 & 324,859 \\
camshape800 & YES & 720 & 232,765 & YES & 884 & 252,936 & YES & 1,316 & 445,334 \\
camshape200 & YES & 900 & 205,779 & YES & 1,413 & 161,270 & YES & 1,598 & 510,171 \\
camshape100 & YES & 467 & 16,756 & YES & 634 & 110,259 & YES & 2,170 & 438,354 \\
btest14 & YES & 365 & 118,417 & YES & 1,038 & 132,136 & YES & 788 & 266,489 \\
beuster & YES & 10,440 & 1,210,906 & NO & - & - & NO & - & - \\
bayes2\_30 & YES & 3,887 & 224,083 & YES & 1,875 & 269,451 & YES & 2,143 & 231,425 \\
bayes2\_20 & YES & 1,816 & 217,604 & YES & 1,068 & 223,733 & YES & 2,314 & 262,980 \\
arki0012 & YES & 1,745 & 417,930 & NO & - & - & YES & 7,335 & 920,327 \\
arki0011 & YES & 703 & 157,341 & YES & 1,247 & 261,525 & YES & 1,550 & 142,736 \\
arki0010 & YES & 508 & 195,330 & YES & 628 & 170,298 & YES & 592 & 302,760 \\
arki009 & YES & 816 & 166,381 & YES & 1,064 & 173,050 & YES & 3,452 & 311,675 \\
arki006 & YES & 3,840 & 2,224,547 & YES & 7,536 & 558,761 & NO & - & - \\
arki004 & YES & 1,080 & 238,489 & YES & 1,717 & 330,403 & NO & - & - \\
arki002 & YES & 404 & 120,740 & YES & 791 & 164,558 & YES & 970 & 370,699 \\
ann\_peaks\_tanh & YES & 608 & 142,221 & YES & 402 & 98,772 & YES & 781 & 149,761 \\
ann\_peaks\_exp & YES & 647 & 122,942 & YES & 802 & 151,447 & NO & - & - \\
ann\_fermentation\_tanh & YES & 632 & 124,303 & YES & 736 & 135,406 & NO & - & - \\
ann\_fermentation\_exp & YES & 548 & 117,099 & YES & 501 & 72,721 & YES & 887 & 127,660 \\
ann\_cumene\_tanh & YES & 527 & 149,772 & YES & 556 & 148,415 & YES & 820 & 287,296 \\
ann\_cumene\_exp & YES & 615 & 206,200 & YES & 664 & 209,323 & YES & 546 & 228,799 \\
ann\_compressor\_tanh & YES & 619 & 121,589 & YES & 561 & 139,337 & YES & 1,072 & 199,087 \\
acopf\_caseactivsg70k\_qcqp & YES & 24,199 & 1,291,624 & NO & - & - & NO & - & - \\
acopf\_caseactivsg25k & YES & 5,460 & 360,159 & NO & - & - & NO & - & - \\
acopf\_case9241pegase\_qcqp & YES & 10,087 & 296,653 & YES & 6,879 & 281,590 & YES & 4,157 & 238,994 \\
acopf\_case6468rte\_qcqp & YES & 3,597 & 285,415 & YES & 4,639 & 239,463 & YES & 10,148 & 547,044 \\
acopf\_case13659pegase\_qcqp & YES & 1,740 & 190,582 & YES & 1,799 & 144,929 & YES & 3,318 & 305,207 \\
acopf\_case1354pegase\_qcqp & YES & 2,254 & 237,694 & YES & 5,053 & 280,002 & YES & 1,687 & 229,763 \\
\bottomrule
\end{tabular*}
\end{table}
\endgroup
\clearpage
\begingroup
\setlength{\intextsep}{5pt}
\begin{table}[H]
\centering
\caption{Per-instance MINLPLib results, panel 2.}
\label{tab:instances-minlplib-1}
\fontsize{8.5}{9}\selectfont
\setlength{\tabcolsep}{2pt}
\renewcommand{\arraystretch}{1}
\begin{tabular*}{\linewidth}{@{\extracolsep{\fill}}l*{2}{crr}@{}}
\toprule
Instance & \multicolumn{3}{c}{AutoEoH} & \multicolumn{3}{c}{EvoX} \\
\cmidrule(lr){2-4}\cmidrule(lr){5-7}
 & Result & Time (s) & Token & Result & Time (s) & Token \\ \midrule
elec100 & NO & - & - & NO & - & - \\
gabriel09 & YES & 383 & 109,927 & YES & 692 & 242,582 \\
etamac & YES & 1,473 & 221,210 & YES & 1,646 & 272,268 \\
eq6\_1 & YES & 557 & 118,939 & YES & 2,508 & 230,076 \\
emfl050 & YES & 626 & 175,207 & YES & 982 & 249,243 \\
csched2 & NO & - & - & NO & - & - \\
eg\_disc2 & YES & 1,158 & 322,742 & NO & - & - \\
edgecross24-115 & NO & - & - & NO & - & - \\
dtoc5 & YES & 1,146 & 261,275 & YES & 992 & 256,388 \\
deb10 & YES & 704 & 142,464 & NO & - & - \\
pooling\_ct3 & NO & - & - & YES & 1,073 & 247,539 \\
crudeoil\_li01 & YES & 6,340 & 439,523 & NO & - & - \\
contvar & YES & 1,114 & 327,686 & YES & 2,276 & 399,203 \\
color\_lab6b\_4x20 & YES & 6,196 & 283,413 & NO & - & - \\
color\_lab3\_4x0 & YES & 8,089 & 303,319 & YES & 3,587 & 282,293 \\
color\_lab2\_4x0 & NO & - & - & YES & 1,081 & 208,158 \\
chp\_shorttermplan2d & NO & - & - & NO & - & - \\
chp\_shorttermplan2b & NO & - & - & NO & - & - \\
chimera\_mgw-c16-2031-01 & NO & - & - & NO & - & - \\
chimera\_mgw-c8-507-onc8-01 & NO & - & - & NO & - & - \\
chimera\_rfr-01 & NO & - & - & YES & 3,518 & 244,440 \\
chimera\_mgw-c16-2031-02 & NO & - & - & NO & - & - \\
chimera\_lga-02 & NO & - & - & NO & - & - \\
chain50 & YES & 864 & 179,070 & YES & 786 & 131,219 \\
chain400 & YES & 1,002 & 204,913 & YES & 684 & 324,151 \\
chain200 & YES & 1,037 & 241,699 & NO & - & - \\
chain100 & YES & 1,047 & 218,353 & NO & - & - \\
cesam2log & YES & 1,681 & 208,622 & YES & 868 & 230,976 \\
cesam2cent & NO & - & - & YES & 1,021 & 234,126 \\
catmix100 & NO & - & - & YES & 650 & 229,948 \\
case\_1scv2 & YES & 1,381 & 184,725 & YES & 876 & 230,808 \\
blendgap & NO & - & - & NO & - & - \\
casctanks & NO & - & - & NO & - & - \\
camshape800 & YES & 802 & 175,577 & YES & 835 & 179,399 \\
camshape200 & NO & - & - & NO & - & - \\
camshape100 & YES & 1,208 & 171,607 & YES & 817 & 208,955 \\
btest14 & YES & 868 & 185,011 & YES & 591 & 183,742 \\
beuster & NO & - & - & NO & - & - \\
bayes2\_30 & NO & - & - & NO & - & - \\
bayes2\_20 & NO & - & - & NO & - & - \\
arki0012 & NO & - & - & YES & 1,175 & 207,922 \\
arki0011 & NO & - & - & YES & 777 & 182,896 \\
arki0010 & YES & 692 & 191,775 & YES & 1,070 & 255,476 \\
arki009 & YES & 5,526 & 1,158,134 & NO & - & - \\
arki006 & NO & - & - & NO & - & - \\
arki004 & YES & 1,034 & 270,011 & NO & - & - \\
arki002 & NO & - & - & YES & 612 & 232,008 \\
ann\_peaks\_tanh & YES & 1,000 & 142,099 & YES & 488 & 179,109 \\
ann\_peaks\_exp & YES & 736 & 133,104 & YES & 616 & 157,326 \\
ann\_fermentation\_tanh & YES & 948 & 84,689 & YES & 558 & 132,529 \\
ann\_fermentation\_exp & YES & 643 & 94,214 & YES & 570 & 79,193 \\
ann\_cumene\_tanh & YES & 713 & 251,374 & YES & 585 & 189,980 \\
ann\_cumene\_exp & YES & 514 & 215,174 & YES & 712 & 243,930 \\
ann\_compressor\_tanh & NO & - & - & YES & 686 & 180,638 \\
acopf\_caseactivsg70k\_qcqp & NO & - & - & NO & - & - \\
acopf\_caseactivsg25k & NO & - & - & NO & - & - \\
acopf\_case9241pegase\_qcqp & NO & - & - & NO & - & - \\
acopf\_case6468rte\_qcqp & NO & - & - & NO & - & - \\
acopf\_case13659pegase\_qcqp & NO & - & - & YES & 1,103 & 134,925 \\
acopf\_case1354pegase\_qcqp & NO & - & - & NO & - & - \\
\bottomrule
\end{tabular*}
\end{table}
\endgroup

\end{document}